\documentclass[11pt]{article}

\usepackage[preprint]{acl}

\usepackage{times}
\usepackage{latexsym}
\usepackage{amsmath} 
\usepackage[ruled,vlined,linesnumbered]{algorithm2e}
\usepackage{amsfonts}

\usepackage[T1]{fontenc}

\usepackage[utf8]{inputenc}

\usepackage{microtype}

\usepackage{inconsolata}

\usepackage{graphicx}

\usepackage{times}
\usepackage{bbm}

\newcommand{\clip}{\mathrm{clip}}

\usepackage{arydshln}
\usepackage{booktabs}



\usepackage{booktabs}
\usepackage{multirow}
\usepackage{arydshln}

\definecolor{SDEblue}{RGB}{28 58 88}
\definecolor{lighterblue}{HTML}{f2fafd}  
\definecolor{cc3}{RGB}{255, 191, 0}
\definecolor{cc4}{RGB}{0, 128, 128}
\hypersetup{
 colorlinks=true,
 linkcolor=SDEblue,   
 urlcolor=magenta,
 citecolor=SDEblue,
}

\usepackage{fontawesome5}
\usepackage[utf8]{inputenc}
\usepackage[T1]{fontenc}
\definecolor{githubblue}{HTML}{5899da}

\hypersetup{
    colorlinks=true,
    urlcolor=githubblue,
}

\title{AdaSR: Adaptive Streaming Reasoning with\\ Hierarchical Relative Policy Optimization}

\author{
 \textbf{Junlong Tong\textsuperscript{1,2}},
 \textbf{Wenqi Xu\textsuperscript{1,5}},
 \textbf{Yingqi Fan\textsuperscript{1}},
 \textbf{Anhao Zhao\textsuperscript{1,3}}\\
 \textbf{Xuan Lu\textsuperscript{1,2}},
 \textbf{Yang Tan\textsuperscript{1,4}},
 \textbf{Xiaoyu Shen\textsuperscript{1}\thanks{Corresponding author}}
\\
 \textsuperscript{1}Eastern Institute of Technology, Ningbo
 \textsuperscript{2}Shanghai Jiao Tong University\\
 \textsuperscript{3}The Hong Kong Polytechnic University
 \textsuperscript{4}Southeast University\\
 \textsuperscript{5}Xi'an Jiaotong-Liverpool University \\
\texttt{jl-tong@sjtu.edu.cn}~~~~~\texttt{xyshen@eitech.edu.cn}
}

\begin{document}
\maketitle

\begin{abstract}
Large reasoning models typically follow a \textit{read-then-think} paradigm: they observe the complete input, reason over a static context, and then produce the answer. Yet many real-world scenarios are inherently dynamic, such as audio and video stream, where information arrives as a continuous stream and models must reason, update, and respond under partial observations. Recent streaming reasoning methods allow models to \textit{think while reading}, but they largely rely on supervised imitation of pre-constructed trajectories, which limits their flexibility. In this paper, we propose \textbf{AdaSR}, an adaptive streaming reasoning framework that enables models to reason during input streaming and perform final deliberation once the stream is complete, learning when to think, and how much computation to allocate across different stages. To optimize this hierarchical reasoning process, we introduce \textbf{Hierarchical Relative Policy Optimization (HRPO)}, which decomposes policy optimization into streaming reasoning and deep reasoning phases, providing more \textit{fine-grained advantage assignment} instead of uniformly distributing a single sequence-level advantage over all tokens. HRPO integrates format, accuracy, and adaptive thinking rewards to enforce valid reasoning protocols, preserve final task performance, and encourage latency-aware computation allocation. Experiments show that AdaSR achieves a better balance among reasoning accuracy, computational efficiency, and streaming latency compared with supervised fine-tuning baseline.
We release our code at \textcolor{githubblue}{\faGithub}\ \href{https://github.com/EIT-NLP/StreamingLLM/tree/main/AdaSR}{\textbf{\texttt{EIT-NLP/AdaSR}}}.
\end{abstract}
\section{Introduction}
\label{sec:1_intro}
Large reasoning models~\citep{openai2024o1,deepseek2025r1} have achieved remarkable performance on complex tasks such as mathematical reasoning, code generation, and multi-step decision making, often through chain-of-thought (CoT) reasoning~\citep{wei2022chain}: they receive an input, generate intermediate reasoning steps, and finally produce an answer. This paradigm follows a read then think pattern, where reasoning begins only after the model has observed a static and complete context. Real-world environments, however, are often dynamic. Inputs do not always appear as complete contexts; instead, they arrive as continuous streams: speech unfolds over time, videos reveal information frame by frame, interactive agents receive observations sequentially, and sensor-driven systems must react before an event is fully observed~\citep{gu2017learning,ma2018stacl,arivazhagan2019monotonic,chen2024videollm,lin2024streamingbench,defossez2024moshi}. In such settings, waiting for the full input introduces unnecessary latency, whereas reasoning too early may lead to premature or misleading conclusions. This raises a fundamental question: how can reasoning models think, update, and respond under continuously evolving observations?

Recent efforts have begun to explore streaming reasoning, enabling models to reason in real time as inputs dynamically unfold~\citep{tong2025streamingthinker,zhang2026tays}. However, existing streaming CoT methods typically rely on supervised fine-tuning over carefully constructed streaming trajectories. Such fine-grained supervision is costly, since each partial observation may require a corresponding local reasoning annotation. More importantly, imitation learning tends to encourage models to reproduce the surface form of streaming traces, rather than acquire the underlying ability to decide whether a partial input requires shallow understanding, deeper reasoning, or no reasoning at all. As a result, models trained on fixed supervised trajectories may exhibit the appearance of streaming reasoning, yet still lack genuine adaptivity.

\begin{figure*}[t]
    \centering
    \includegraphics[width=1\linewidth]{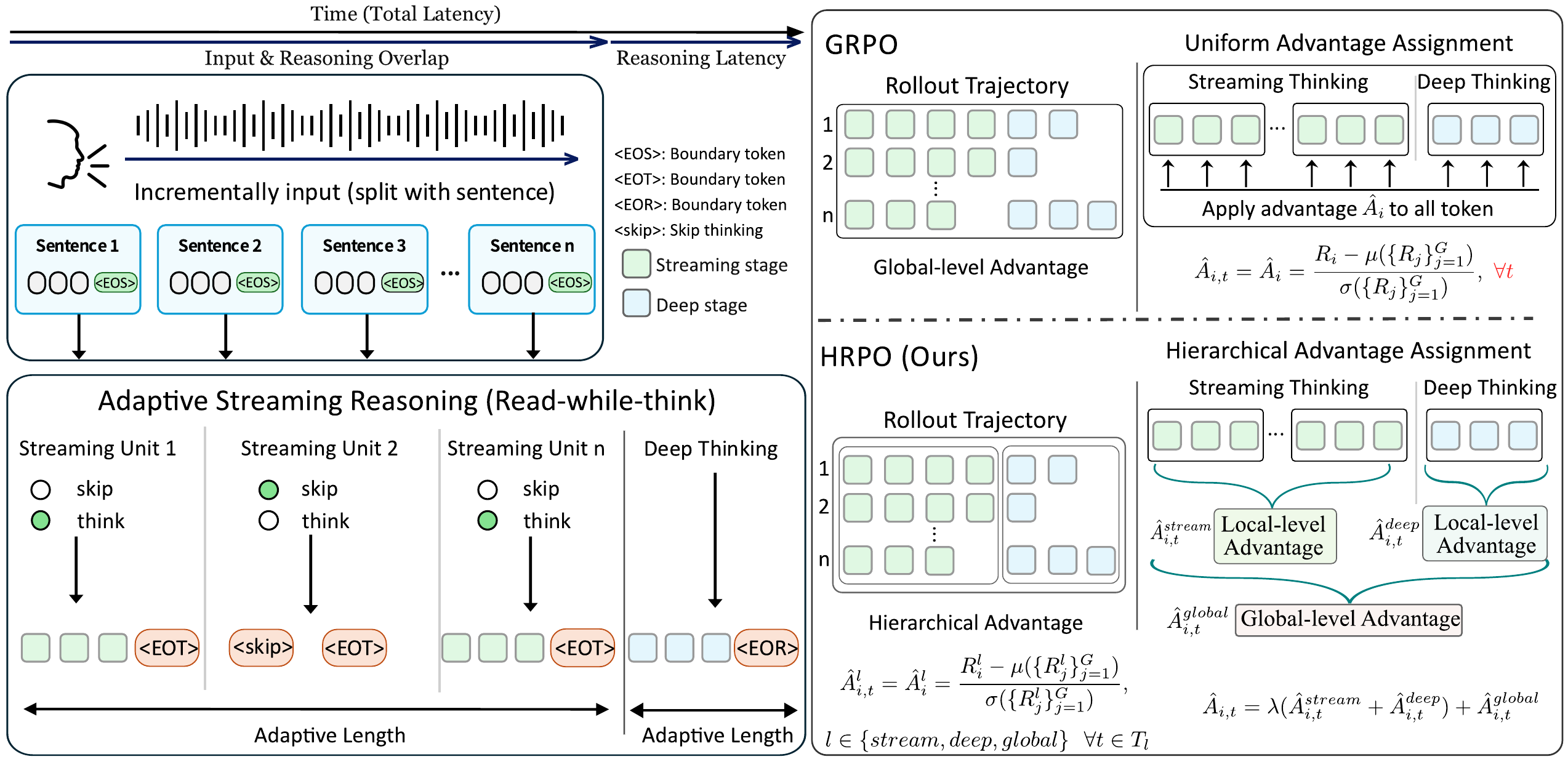}
    \caption{Overview of the AdaSR framework. \textit{(left)} AdaSR thinks while reading streaming inputs, decide whether to reason over each segment, and allocate computation between streaming thinking and final deep thinking. \textit{(right)} AdaSR introduces HRPO to align optimization with the temporal structure of streaming reasoning, assigning fine-grained advantages across local and global scopes.}
    \label{fig:framework}
    \vspace{-1.5em}
\end{figure*}

To move beyond imitation-based streaming reasoning, we propose AdaSR, an adaptive streaming reasoning framework that leverages reinforcement learning to optimize how models reason over evolving inputs, as shown in Figure~\ref{fig:framework}. AdaSR learns a computation policy during input streaming: when to reason, when to skip, and how to balance streaming reasoning with final deliberation, echoing broader efforts on adaptive computation and test-time compute allocation~\citep{graves2016adaptive,snell2024scaling,muennighoff2025s1}. A natural starting point is Group Relative Policy Optimization (GRPO)~\citep{shao2024deepseekmath}, which has shown strong effectiveness in improving reasoning models with outcome-based rewards~\citep{ziegler2019fine,ouyang2022training,cobbe2021training,shao2024deepseekmath,deepseek2025r1}. However, streaming reasoning presents a structured optimization problem that standard GRPO does not explicitly capture. A streaming trajectory consists of multiple reasoning units generated under partial observations, followed by a deep reasoning phase that integrates the complete context. Assigning a single sequence-level advantage uniformly to all generated tokens blurs this temporal structure, making it difficult to distinguish the contributions of local understanding, global integration, and final correctness.\footnote{
GRPO's uniform advantage assignment leads to a cross-phase credit paradox: deep-phase success can falsely amplify redundant streaming thoughts, while deep-phase inefficiency can unfairly suppress useful ones.
}

To address this issue, AdaSR introduces Hierarchical Relative Policy Optimization (HRPO), which refines GRPO with stage-aware advantage assignment. Instead of treating the entire trajectory as a flat sequence, HRPO decomposes the learning signal according to the temporal structure of streaming reasoning, assigning distinct advantages to streaming tokens, deep-reasoning tokens, and the full rollout. This preserves GRPO’s group-relative optimization while turning the coarse sequence-level advantage into finer-grained, hierarchy-aware credit signals. Combined with format, accuracy, and adaptive thinking rewards, AdaSR optimizes both final task performance and the computation path that leads to it. Experiments show that AdaSR achieves a better balance among reasoning accuracy, computational efficiency, and latency compared with supervised fine-tuning baselines.

Our contributions are fourfold:
\textbf{(1)} We propose AdaSR, to the best of our knowledge the first adaptive streaming reasoning framework that learns when to reason, when to skip, and how to allocate computation over evolving inputs, guided by adaptive thinking rewards.
\textbf{(2)} We introduce HRPO, a hierarchy-aware advantage assignment method that replaces coarse sequence-level credit with local-to-global advantages, enabling fine-grained optimization for structured streaming reasoning trajectories.
\textbf{(3)} We adapt streaming reasoning to a vLLM-compatible rollout and inference pipeline~\citep{kwon2023efficient}, improving its practicality for efficient training and deployment in realistic streaming scenarios.
\textbf{(4)} Experiments show that AdaSR achieves a better trade-off among reasoning accuracy and latency than supervised fine-tuning baselines.

\section{Preliminary}
\paragraph{Streaming Thinking}

The standard reasoning paradigm follows a \emph{read-then-think} pattern: the model receives the full context before reasoning. Given a question $Q$ and a context $C=\{C_1,\ldots,C_T\}$ decomposed into sequential sentences, with $R_t$ denoting the reasoning segment associated with the prefix ending at $C_t$, this component can be summarized as \(\mathcal{P}_{\mathrm{standard}}=\textstyle\prod_{t=1}^{T}P(R_t|Q,C_{\leq T},R_{\leq t-1})\).
In contrast, \emph{streaming thinking}~\citep{tong2025streamingthinker,zhang2026tays,chiang2025shanks} lets reasoning unfold with the input stream: as each sentence arrives, the model forms a local thought under partial context, then performs a final deep deliberation once the complete context and task instruction $I$ are available:
\begin{equation}
\label{eq:paradigm}
\resizebox{\columnwidth}{!}{$\displaystyle
\begin{aligned}
\mathcal{P}_{\mathrm{streaming}}
&=
P(R_q \mid Q)
\prod_{t=1}^{T}
P(R_t \mid Q, C_{\leq t}, R_{\leq t-1}) \\
&\quad\cdot
P(R \mid Q, C_{\leq T}, R_{\leq T}) .
\end{aligned}
$}
\end{equation}
The first factor is the streaming phase: after each sentence $C_t$, the model emits $R_t$ before seeing future sentences, or skips irrelevant input. Boundary and end-of-thought tokens separate local segments under streaming masks. The second factor is the deep phase, integrating $C_{\leq T}$ and $R_{\leq T}$ into the final response $R$. We denote the phase boundary by $t_{\mathrm{s}}$.
\paragraph{GRPO for Reasoning}

Reinforcement learning improves LLM reasoning by optimizing outcome rewards without explicit chain-of-thought annotations~\citep{deepseek2025r1}. In PPO~\citep{schulman2017proximal}, old-policy rollouts are reweighted token by token and clipped to stabilize updates, while advantages are typically estimated by a learned value function. GRPO~\citep{shao2024deepseekmath} keeps this clipped update but removes the value model: for a question-answer pair $(q,a)\sim\mathcal{D}$, the old policy $\pi_{\theta_{\mathrm{old}}}$ samples $G$ candidate outputs $\{o_i\}_{i=1}^{G}$, whose rewards are compared within the same group to form relative advantages. The objective averages the clipped token-level surrogate over each candidate sequence while keeping $\pi_\theta$ close to $\pi_{\mathrm{ref}}$:
\begin{equation}
\label{eq:grpo}
\begin{aligned}
&J_{\mathrm{GRPO}}(\theta)
=
\mathbb{E}_{\substack{
(q,a) \sim \mathcal{D} \\
\{o_i\}_{i=1}^{G}
\sim \pi_{\theta_{\mathrm{old}}}(\cdot \mid q)
}}
\Bigg\{
\frac{1}{G}\sum_{i=1}^{G}
\Bigg[
\\
&\quad
\frac{1}{|o_i|}
\sum_{t=1}^{|o_i|}
\mathcal{C}\!\left(r_{i,t}(\theta), \hat{A}_i\right)
-
\beta\,
\operatorname{KL}\!\left(
\pi_{\theta} \,\|\, \pi_{\mathrm{ref}}
\right)
\Bigg]
\Bigg\},
\end{aligned}
\end{equation}
where the clipped term \(\mathcal{C}\big(r_{i,t}(\theta),\hat{A}_i\big)=\min\!\big(r_{i,t}(\theta)\hat{A}_i,clip(r_{i,t}(\theta),1-\varepsilon,1+\varepsilon)\hat{A}_i\big)\) inherits PPO's trust-region behavior through the clipping threshold $\varepsilon$, with $\beta$ controlling the KL regularization strength. In this term, the token-wise importance ratio is paired with a sequence-level advantage obtained by normalizing the reward $R_i=R(q,a,o_i)$ against the $G$ rewards sampled for the same question:
\begin{equation}
\label{eq:grpo_ratio_adv}
\resizebox{\columnwidth}{!}{$\displaystyle
\begin{aligned}
r_{i,t}(\theta)
&=
\frac{
\pi_{\theta}(o_{i,t} \mid q,o_{i,<t})
}{
\pi_{\theta_{\mathrm{old}}}(o_{i,t} \mid q,o_{i,<t})
},
\hat{A}_{i}
=
\frac{
R_i-\mu(\{R_j\}_{j=1}^G)
}{
\sigma(\{R_j\}_{j=1}^G)
}.
\end{aligned}
$}
\end{equation}

Since the same $\hat{A}_i$ is attached to every token position $t$ in $o_i$, GRPO treats a candidate output as one flat trajectory. This is suitable for batch reasoning, where the model reasons after observing the full input, but it is too coarse for streaming reasoning: local thoughts produced under partial observations and final deliberation over the complete context receive indistinguishable credit.

\section{Methodology}
\label{sec:3_method}

We present Adaptive Streaming Reasoning(\textbf{AdaSR}), an RL framework for adaptive computation allocation between streaming and deep reasoning. As shown in Figure~\ref{fig:framework}, AdaSR combines hierarchical relative policy optimization (HRPO \S\ref{sec:hrpo}), adaptive rewards (\S\ref{sec:reward}), and streaming rollout (\S\ref{sec:training}).

\subsection{Hierarchical Relative Policy Optimization}
\label{sec:hrpo}

The GRPO assigns a uniform advantage to all tokens in a trajectory, treating it as a flat sequence. However, streaming reasoning trajectories have an inherently \emph{hierarchical} structure: the streaming phase ($1 \leq t \leq |t_{\text{s}}|$) produces local and partial observations, while the deep phase ($|t_{\text{s}}| < t \leq |o_i|$) integrates the full context. The \textbf{HRPO} decomposes advantage estimation into fine-grained levels.

\paragraph{Hierarchical Advantage Assignment}

The central difficulty in streaming reasoning is temporal credit assignment: streaming tokens should be credited for local decisions under partial observations, deep tokens for final integration, and the whole trajectory for answer correctness. HRPO keeps the group-relative normalization of GRPO, but separates these signals into streaming-local ($s$), deep-local ($d$), and trajectory-global ($g$) advantages.

For each question $q$, HRPO samples $G$ trajectories $\{o_i\}_{i=1}^G$ from $\pi_{\theta_{\mathrm{old}}}$. Let $A_i^s$, $A_i^d$, and $A_i^g$ denote the group-relative advantages associated with the streaming-local, deep-local, and trajectory-global levels, respectively. HRPO defines hierarchical advantages by attaching the appropriate level-wise advantage to the corresponding token range:
\begin{equation}
\label{eq:advantage}
\begin{aligned}
\hat{A}_{i,t}^{\ell}
&=
A_i^{\ell},
\quad t \in \mathcal{T}_i^\ell,
\quad \ell \in \{s,d,g\},
\\
\mathcal{T}_i^s
&= [1, |t_{\mathrm{s}}|],
\mathcal{T}_i^d = (|t_{\mathrm{s}}|, |o_i|],
\mathcal{T}_i^g = [1, |o_i|].
\end{aligned}
\end{equation}
where the intervals in $\mathcal{T}_i^\ell$ denote integer token positions. Here $|t_{\text{s}}|$ denotes the token length of the streaming reasoning phase, so $|o_i|\!-\!|t_{\text{s}}|$ is the length of the deep phase. For each level $\ell\in\{s,d,g\}$, HRPO uses the shared per-token importance ratio \(r_{i,t}^{\ell}(\theta)=\pi_\theta(o_{i,t}|q,o_{i,<t})/\pi_{\theta_{\mathrm{old}}}(o_{i,t}|q,o_{i,<t})\). The superscript records which hierarchical advantage and surrogate term the ratio is paired with. Thus, local advantages shape phase-specific behavior, while the global advantage preserves answer-level correctness across the full rollout.

\paragraph{Policy Optimization Objective}

Based on the fine-grained advantages and ratios above, HRPO defines three clipped surrogate losses over their corresponding token ranges. The objective is:
\begin{equation}
\label{eq:hrpo}
\begin{aligned}
&J_{\mathrm{HRPO}}(\theta)
=
\mathbb{E}_{\substack{
(q,a) \sim \mathcal{D} \\
\{o_i\}_{i=1}^{G} \sim \pi_{\theta_{\mathrm{old}}}
}}
\Bigg\{
\frac{1}{G}\sum_{i=1}^{G}
\Bigg[
\\
&\resizebox{\columnwidth}{!}{$\displaystyle
\underbrace{
\frac{\lambda}{|t_{\mathrm{s}}|}
\sum_{t=1}^{|t_{\mathrm{s}}|}
\mathcal{C}(r_{i,t}^{s}, \hat{A}_{i,t}^{s})
+
\frac{\lambda}{|o_i|-|t_{\mathrm{s}}|}
\sum_{t=|t_{\mathrm{s}}|+1}^{|o_i|}
\mathcal{C}(r_{i,t}^{d}, \hat{A}_{i,t}^{d})
}_{\substack{\text{local clipped surrogate objective}}}
$}
\\
&\resizebox{\columnwidth}{!}{$\displaystyle
+
\underbrace{
\frac{1}{|o_i|}
\sum_{t=1}^{|o_i|}
\Bigg(
\mathcal{C}(r_{i,t}^{g}, \hat{A}_{i,t}^{g})
}_{\substack{\text{global clipped surrogate objective}}}
-
\beta\operatorname{KL}\!\left(
\pi_{\theta} \,\|\, \pi_{\mathrm{ref}}
\right)
\Bigg)
\Bigg]
\Bigg\},
$}
\end{aligned}
\end{equation}
where $\mathcal{C}(r_{i,t}^{\ell}, \hat{A}_{i,t}^{\ell}) = \min\!\big(r_{i,t}^{\ell}\,\hat{A}_{i,t}^{\ell},\;\clip(r_{i,t}^{\ell}, $
$1{-}\varepsilon, 1{+}\varepsilon)\,\hat{A}_{i,t}^{\ell}\big)$ for $\ell\in\{s,d,g\}$ is the clipped surrogate objective~\citep{schulman2017proximal}.\footnote{We provide the policy gradient analysis in Appendix~\ref{app:hrpo-gradient}.}

The coefficient $\lambda \in [0,1]$ balances local and global objectives. When $\lambda = 0$, HRPO falls back to the global trajectory-level signal; increasing $\lambda$ strengthens phase-specific optimization. Unlike the GRPO which applies one advantage uniformly across the entire trajectory, HRPO applies differentiated local advantages to each phase while simultaneously maintaining a global correctness signal across all tokens, resolving the temporal credit assignment problem inherent in streaming reasoning.

\subsection{Adaptive Streaming Reasoning Reward}
\label{sec:reward}

In streaming reasoning, the model must reason incrementally as the input unfolds, deciding not only what intermediate thoughts to produce but also how much computation to allocate at each stage. Optimizing only for terminal accuracy provides no direct supervision over this computation path, and may therefore encourage inefficient trajectories, such as overly verbose streaming thoughts or delayed reasoning concentrated in the deep phase. To guide both task performance and computation allocation, we decompose the reward into format, accuracy, and adaptive thinking components, corresponding to structural validity, final correctness, and latency-aware reasoning efficiency.

\paragraph{Format Reward}

The format reward makes rollouts parseable for HRPO. During streaming, each segment must terminate with \texttt{<EOT>}, and the content before \texttt{<EOT>} must be either a reasoning thought or \texttt{<skip>}. During deep reasoning, the output must contain a non-empty deep reasoning field ending with \texttt{<EOR>}. For rollout $i$, we denote the corresponding binary rewards by $R_{i,\mathrm{fmt}}^s$ and $R_{i,\mathrm{fmt}}^d$. This component constrains structural validity rather than reasoning quality.

\paragraph{Accuracy Reward}

The accuracy reward provides the terminal task signal:
$R_i^{\mathrm{acc}}=\mathbbm{1}[\hat{a}_i=a_i]$, where $\hat{a}_i$ is the model prediction and $a_i$ is the ground-truth answer. It evaluates the complete trajectory after both streaming and deep reasoning, anchoring learning to final correctness.

\paragraph{Adaptive Thinking Reward}

The adaptive thinking reward uses token length as a proxy for computation allocation. It discourages excessive computation while allowing the policy to allocate more tokens when additional reasoning is useful. We first define phase-local length-shaping penalties:
\vspace{-0.5em}
\begin{equation}
\begin{aligned}
R_i^{L_s}
 &=
\frac{1}{N}\sum_{n=1}^{N}
-\log\!\left(1+|s_{i,n}|\right),
\\
R_i^{L_d}
&=
-\log\!\left(1+L_i^D\right),
\end{aligned}
\end{equation}
where $R_i^{L_s}$ and $R_i^{L_d}$ are local length penalties for streaming and deep reasoning, respectively.
These terms are gated by accuracy and format, so length shaping is applied only to correct and parseable trajectories. The logarithmic form discourages verbosity with diminishing marginal strength.

Local penalties alone do not capture the latency asymmetry between phases: streaming reasoning can overlap with input arrival, whereas deep reasoning begins only after the stream ends. We therefore introduce a success-conditioned trajectory-level efficiency reward:
\vspace{-0.5em}
\begin{equation}
\label{eq:reward_global}
\begin{aligned}
&R_i^{\mathrm{eff}}
=
\underbrace{
R_i^{\mathrm{acc}}
R_{i,\mathrm{fmt}}^s
R_{i,\mathrm{fmt}}^d
}_{\text{acc.\ \& format gate}}
\\
&\cdot
\underbrace{
\left(1-\exp\!\left(-\frac{L_i^D}{\tau}\right)\right)
\left(1-\exp\!\left(-\frac{L_i^S}{\tau}\right)\right)
}_{\substack{\text{reasoning length lower-bound gate}}}
\\
&\cdot
\underbrace{
\exp\!\left(
-\frac{\alpha L_i^S+L_i^D}{\tau}
\right)
}_{\text{latency-aware allocation}},
\end{aligned}
\end{equation}
where $0<\alpha<1$ and $\tau>0$ control the latency discount and shared reward scale, respectively.\footnote{$\alpha$ discounts streaming tokens because they can overlap with input arrival, while deep tokens incur full post-stream latency. $\tau$ is the shared scale for both lower-bound gates and the latency penalty. We provide the details in Appendix~\ref{app:beta-sensitivity}.} The lower-bound gates assign little efficiency bonus to trajectories with negligible reasoning in either phase and saturate once sufficient reasoning is produced. The exponential term then penalizes the effective latency cost $\alpha L_i^S+L_i^D$. Thus, the global reward favors trajectories that are correct, parseable, and computationally sufficient but not excessive.

\begin{algorithm*}[!t]
\caption{AdaSR Training with HRPO}
\label{alg:adasr}
\SetAlgoLined
\KwIn{Dataset $\mathcal{D}$, policy $\pi_\theta$ initialized from SFT model, group size $G$, hierarchy weight $\lambda$, clip range $\varepsilon$.}
\KwOut{Trained policy $\pi_\theta$.}
\For{each training iteration}{
    Sample mini-batch $\mathcal{B} = \{(q_k, a_k)\}$ from $\mathcal{D}$\;

    \For{each $(q, a) \in \mathcal{B}$}{
        Split context into sentences $\{C_1, \ldots, C_T\}$\;

        \For(\tcp*[f]{Group sampling}){$i = 1$ \KwTo $G$}{
            \For(\tcp*[f]{Streaming rollout}){$t = 1$ \KwTo $T$}{
                Prefill $C_t$ with streaming position encoding and attention mask (See Appendix~\ref{app:training-details});
                Decode $R_t^{(i)} \sim \pi_{\theta_{\mathrm{old}}}
                (\cdot \mid Q, C_{\leq t}, R_{<t}^{(i)})$ until \texttt{<EOT>}\;
            }

            Decode deep reasoning $R^{(i)} \sim \pi_{\theta_{\mathrm{old}}}
            (\cdot \mid Q, C_{\leq T}, R_{\leq T}^{(i)})$\;

            Collect $o_i = [R_1^{(i)}, \ldots, R_T^{(i)}, R^{(i)}]$ with log-probs\;
        }

        Compute reward components for each $o_i$\;
        Compose $A_i^s$, $A_i^d$, and $A_i^g$ via Eq.~\ref{eq:advantage_composition}\;
        Assign hierarchical token advantages via Eq.~\ref{eq:advantage}\;
        Compute $J_{\mathrm{HRPO}}$ via Eq.~\ref{eq:hrpo}, using the time-grouped form in Eq.~\ref{eq:hrpo_time_grouped}\;
    }

    Update $\theta$ via gradient ascent on $J_{\mathrm{HRPO}}$\;
}
\end{algorithm*}
\paragraph{Hierarchical Advantage Composition}

Eq.~\ref{eq:advantage} specifies where each level-wise advantage is applied: 
$A_i^s$ is assigned to streaming tokens, $A_i^d$ to deep-reasoning tokens, and $A_i^g$ to the full trajectory. 
Following the multi-reward normalization observation in GDPO~\citep{liu2026gdpo}, we avoid first summing weighted rewards and then normalizing the result, which can blur distinct reward components into the same advantage signal. 
Instead, for any component reward $X_i$ within the $G$ rollouts of the same question, we first compute the component advantage $\mathcal{N}_{\mathcal{G}}(X_i)=\big(X_i-\mu(\{X_j\}_{j=1}^G)\big)/\sigma(\{X_j\}_{j=1}^G)$, and then compose advantages at their corresponding temporal levels:
\vspace{-0.5em}
{
\begin{equation}
\label{eq:advantage_composition}
\begin{aligned}
A_i^s
&=
\mathcal{N}_{\mathcal{G}}\!\left(R_{i,\mathrm{fmt}}^s\right)
+
\beta\mathcal{N}_{\mathcal{G}}\!\left(R_i^{L_s}\right),
\\
A_i^d
&=
\mathcal{N}_{\mathcal{G}}\!\left(R_{i,\mathrm{fmt}}^d\right)
+
\beta\mathcal{N}_{\mathcal{G}}\!\left(R_i^{L_d}\right),
\\
A_i^g
&=
\mathcal{N}_{\mathcal{G}}\!\left(R_i^{\mathrm{acc}}\right)
+
\beta\mathcal{N}_{\mathcal{G}}\!\left(R_i^{\mathrm{eff}}\right).
\end{aligned}
\end{equation}
}
This advantage-level composition matches Eq.~\ref{eq:advantage}: format and length signals supervise the corresponding streaming or deep time span, while accuracy and efficiency provide a trajectory-level signal. 
The shared $\beta$ controls the strength of the adaptive thinking signal after normalization. 
\footnote{If $\beta$ were instead applied at the reward level before group normalization, then in the active gated regime the positive scale would be canceled and $\beta$ would lose its effect; we provide the proof in Appendix~\ref{app:beta-sensitivity}.}

\subsection{Training Algorithm}
\label{sec:training}

\paragraph{Streaming Rollout}

Following StreamingThinker~\citep{tong2025streamingthinker}, we implement streaming rollout by extending the vLLM inference engine~\citep{kwon2023efficient}. Input sentences are fed sequentially: after prefilling each sentence $C_t$, the model decodes streaming reasoning tokens $R_t$ until \texttt{<EOT>}. Position IDs implement streaming position encoding---input tokens and reasoning tokens maintain independent positional indices~\citep{su2024roformer, tong2025streamingthinker}. Streaming attention masks prevent attending to future input during the streaming phase. After all $T$ sentences, the instruction $I$ is appended and the model generates deep reasoning $R$. The trajectory $o_i = [R_1, \ldots, R_T, R]$ is collected with per-token log-probabilities.

\paragraph{Policy Gradient Computation}

Given the collected trajectories, we compute log-probabilities under $\pi_\theta$ using streaming attention masks. The reward components are computed, converted into group-relative component advantages, composed into level-wise advantages, and assigned to token ranges before optimizing the HRPO loss (Eq.~\ref{eq:hrpo}) via gradient ascent.
\footnote{Appendix~\ref{app:training-details} provides implementation details on the streaming vLLM rollout, attention masks, stream-aware logits computation, veRL adaptation, and the full algorithm.}
The training process is shown in the Algorithm~\ref{alg:adasr}.

\section{Experiments}
\paragraph{Experimental Setup}
\label{subsec:exp_setup}

We evaluate AdaSR with Qwen3 series models~\citep{yang2025qwen3} on reasoning tasks covering mathematical reasoning, context-based question answering, and logical reasoning. For in-domain evaluation, we use GSM-Symbolic~\citep{mirzadeh2024gsmsymbolic}, MetaMathQA~\citep{yu2023metamath}, and PubMedQA~\citep{jin2019pubmedqa}; for out-of-domain evaluation, we further use GSM-Infinite~\citep{zhou2025gsminfinite} and LogicNLI dataset~\citep{tian-etal-2021-diagnosing}. These datasets contain sufficiently long problem statements, contextual passages, or multi-step constraints that can be naturally segmented into sentence-level streams, making them suitable for evaluating reasoning under partial and progressively arriving inputs.

We evaluate AdaSR in terms of reasoning performance and streaming efficiency. Accuracy is the primary metric, reflecting the goal of improving reasoning under streaming inputs. We further measure efficiency using token length and real latency.\footnote{See Appendix~\ref{subsec:latency} for the definition of real latency.} Token length includes total generation length as well as streaming- and deep-reasoning lengths, while latency captures response delay under streaming inference, where streaming reasoning can overlap with input reception but deep reasoning directly delays the final answer.

\paragraph{Main Experiment}
\begin{table*}[t]
\centering
\scriptsize
\setlength{\tabcolsep}{4.0pt}
\renewcommand{\arraystretch}{0.86}
\definecolor{sftgain}{HTML}{D00000}
\definecolor{sftloss}{HTML}{087F23}
\newcommand{\sftpos}[1]{{\normalfont\tiny\textcolor{sftgain}{(#1\%)}}}
\newcommand{\sftneg}[1]{{\normalfont\tiny\textcolor{sftloss}{(#1\%)}}}
\resizebox{\textwidth}{!}{%
\begin{tabular}{lllccr@{\hspace{0.25em}}lr@{\hspace{0.25em}}l}
\toprule
Model & Benchmark & Metric
& Read-then-think
& StreamingThinker (SFT)~\citep{tong2025streamingthinker}
& \multicolumn{2}{c}{AdaSR-GRPO (Ours)}
& \multicolumn{2}{c}{AdaSR-HRPO (Ours)} \\
\midrule

\multirow{16}{*}{\rotatebox[origin=c]{90}{\textit{Qwen3-1.7B}}}
& \multirow{4}{*}{GSM-symbolic P1}
& Acc$\uparrow$ & 0.634 & 0.834 & 0.850 & \sftpos{+1.9} & \textbf{0.871} & \sftpos{+4.4} \\
& & Streaming stage Len.$\downarrow$ & -- & 165.707 & 172.994 & \sftneg{+4.4} & \textbf{160.032} & \sftpos{-3.4} \\
& & Deep stage Len.$\downarrow$ & 1598.699 & \textbf{118.042} & 133.798 & \sftneg{+13.3} & 131.972 & \sftneg{+11.8} \\
& & Total Len.$\downarrow$ & 1598.699 & \textbf{283.749} & 306.792 & \sftneg{+8.1} & 292.004 & \sftneg{+2.9} \\
\cmidrule(lr){2-9}

& \multirow{4}{*}{GSM-symbolic P2}
& Acc$\uparrow$ & 0.424 & 0.642 & 0.758 & \sftpos{+18.1} & \textbf{0.788} & \sftpos{+22.7} \\
& & Streaming stage Len.$\downarrow$ & -- & 229.290 & 235.784 & \sftneg{+2.8} & \textbf{210.046} & \sftpos{-8.4} \\
& & Deep stage Len.$\downarrow$ & 1866.474 & \textbf{132.616} & 148.704 & \sftneg{+12.1} & 160.210 & \sftneg{+20.8} \\
& & Total Len.$\downarrow$ & 1866.474 & \textbf{361.906} & 384.488 & \sftneg{+6.2} & 370.256 & \sftneg{+2.3} \\
\cmidrule(lr){2-9}

& \multirow{4}{*}{MetaMathQA}
& Acc$\uparrow$ & 0.848 & 0.688 & 0.823 & \sftpos{+19.6} & \textbf{0.826} & \sftpos{+20.1} \\
& & Streaming stage Len.$\downarrow$ & -- & 149.196 & 158.712 & \sftneg{+6.4} & 173.246 & \sftneg{+16.1} \\
& & Deep stage Len.$\downarrow$ & 1281.774 & 121.465 & 159.285 & \sftneg{+31.1} & \textbf{129.356} & \sftneg{+6.5} \\
& & Total Len.$\downarrow$ & 1281.774 & 270.661 & 317.997 & \sftneg{+17.5} & \textbf{302.602} & \sftneg{+11.8} \\
\cmidrule(lr){2-9}

& \multirow{4}{*}{PubMedQA}
& Acc$\uparrow$ & 0.588 & 0.749 & 0.776 & \sftpos{+3.6} & \textbf{0.793} & \sftpos{+5.9} \\
& & Streaming stage Len.$\downarrow$ & -- & 414.737 & 433.573 & \sftneg{+4.5} & \textbf{414.206} & \sftpos{-0.1} \\
& & Deep stage Len.$\downarrow$ & 434.629 & \textbf{120.534} & 143.314 & \sftneg{+18.9} & 136.642 & \sftneg{+13.4} \\
& & Total Len.$\downarrow$ & 434.629 & 535.271 & 576.887 & \sftneg{+7.8} & \textbf{550.848} & \sftneg{+2.9} \\

\midrule

\multirow{16}{*}{\rotatebox[origin=c]{90}{\textit{Qwen3-4B}}}
& \multirow{4}{*}{GSM-symbolic P1}
& Acc$\uparrow$ & 0.636 & 0.934 & 0.946 & \sftpos{+1.3} & \textbf{0.950} & \sftpos{+1.7} \\
& & Streaming stage Len.$\downarrow$ & -- & 167.786 & 157.305 & \sftpos{-6.2} & \textbf{150.688} & \sftpos{-10.2} \\
& & Deep stage Len.$\downarrow$ & 1717.155 & 119.206 & \textbf{111.145} & \sftpos{-6.8} & 116.559 & \sftpos{-2.2} \\
& & Total Len.$\downarrow$ & 1717.155 & 286.992 & 268.450 & \sftpos{-6.5} & \textbf{267.247} & \sftpos{-6.9} \\
\cmidrule(lr){2-9}

& \multirow{4}{*}{GSM-symbolic P2}
& Acc$\uparrow$ & 0.462 & 0.816 & 0.874 & \sftpos{+7.1} & \textbf{0.884} & \sftpos{+8.3} \\
& & Streaming stage Len.$\downarrow$ & -- & 227.210 & 209.866 & \sftpos{-7.6} & \textbf{196.390} & \sftpos{-13.6} \\
& & Deep stage Len.$\downarrow$ & 1937.396 & 130.374 & \textbf{126.268} & \sftpos{-3.1} & 131.162 & \sftneg{+0.6} \\
& & Total Len.$\downarrow$ & 1937.396 & 357.584 & 336.134 & \sftpos{-6.0} & \textbf{327.552} & \sftpos{-8.4} \\
\cmidrule(lr){2-9}

& \multirow{4}{*}{MetaMathQA}
& Acc$\uparrow$ & 0.828 & 0.860 & 0.909 & \sftpos{+5.7} & \textbf{0.924} & \sftpos{+7.4} \\
& & Streaming stage Len.$\downarrow$ & -- & 152.200 & 140.010 & \sftpos{-8.0} & \textbf{121.708} & \sftpos{-20.0} \\
& & Deep stage Len.$\downarrow$ & 1379.088 & 121.998 & \textbf{113.545} & \sftpos{-6.9} & 115.650 & \sftpos{-5.2} \\
& & Total Len.$\downarrow$ & 1379.088 & 274.198 & 253.555 & \sftpos{-7.5} & \textbf{237.358} & \sftpos{-13.4} \\
\cmidrule(lr){2-9}

& \multirow{4}{*}{PubMedQA}
& Acc$\uparrow$ & 0.615 & 0.756 & 0.792 & \sftpos{+4.8} & \textbf{0.807} & \sftpos{+6.7} \\
& & Streaming stage Len.$\downarrow$ & -- & 402.877 & 378.214 & \sftpos{-6.1} & \textbf{356.728} & \sftpos{-11.5} \\
& & Deep stage Len.$\downarrow$ & 331.835 & 116.106 & \textbf{111.842} & \sftpos{-3.7} & 114.512 & \sftpos{-1.4} \\
& & Total Len.$\downarrow$ & 331.835 & 518.983 & 490.056 & \sftpos{-5.6} & \textbf{471.240} & \sftpos{-9.2} \\

\bottomrule
\end{tabular}
}
\caption{Main results on streaming reasoning benchmarks with Qwen3-1.7B/4B. We compare AdaSR-HRPO with GRPO, read-then-think, and SFT-based baselines, reporting accuracy and streaming/deep/total lengths. Percentages denote relative changes over StreamingThinker (SFT); red indicates improvements and green degradations.}
\label{tab:main_results}
\end{table*}

\begin{table*}[t]
\centering
\small
\setlength{\tabcolsep}{5.0pt}
\renewcommand{\arraystretch}{0.92}
\definecolor{sftgain}{HTML}{D00000}
\definecolor{sftloss}{HTML}{087F23}
\newcommand{\sftpos}[1]{{\normalfont\tiny\textcolor{sftgain}{(#1\%)}}}
\newcommand{\sftneg}[1]{{\normalfont\tiny\textcolor{sftloss}{(#1\%)}}}
\resizebox{\textwidth}{!}{%
\begin{tabular}{lcccccccc}
\toprule
Qwen3-1.7B
& \multicolumn{4}{c}{GSM-symbolic P2}
& \multicolumn{4}{c}{GSM-symbolic P1} \\
\cmidrule(lr){2-5}
\cmidrule(lr){6-9}
Method
& Acc$\uparrow$ & Streaming Len.$\downarrow$ & Deep Len.$\downarrow$ & Total Len.$\downarrow$
& Acc$\uparrow$ & Streaming Len.$\downarrow$ & Deep Len.$\downarrow$ & Total Len.$\downarrow$ \\
\midrule
GRPO
& 0.758 & 235.784 & 148.704 & 384.488
& 0.867 & 172.994 & 133.798 & 306.792 \\
\noalign{\vskip 1pt}
\cdashline{1-9}
\noalign{\vskip 1pt}
HRPO
& \textbf{0.788} \sftpos{+4.0} & \textbf{210.046} \sftpos{-10.9} & 160.210 \sftneg{+7.7} & 370.256 \sftpos{-3.7}
& \textbf{0.871} \sftpos{+0.5} & 160.032 \sftpos{-7.5} & 131.972 \sftpos{-1.4} & 292.004 \sftpos{-4.8} \\
HRPO-sentence
& 0.754 \sftneg{-0.5} & 250.946 \sftneg{+6.4} & \textbf{131.024} \sftpos{-11.9} & 381.970 \sftpos{-0.7}
& 0.868 \sftpos{+0.1} & 180.677 \sftneg{+4.4} & \textbf{115.770} \sftpos{-13.5} & 296.447 \sftpos{-3.4} \\
HRPO-token
& 0.770 \sftpos{+1.6} & 213.528 \sftpos{-9.4} & 145.432 \sftpos{-2.2} & \textbf{358.960} \sftpos{-6.6}
& 0.862 \sftneg{-0.6} & \textbf{158.733} \sftpos{-8.2} & 120.714 \sftpos{-9.8} & \textbf{279.447} \sftpos{-8.9} \\
\bottomrule
\end{tabular}
}
\caption{Effect of hierarchical advantage assignment on GSM-symbolic P2/P1. We report accuracy and phase-specific lengths for GRPO, HRPO, sentence-level, and token-level variants. Percentages are relative to GRPO; red/green denote improvements/degradations. Token-level assignment only affects the format boundary token.}
\label{tab:hierarchical_advantage}
\end{table*}

\begin{figure*}[t]
    \centering
    \includegraphics[width=1\linewidth]{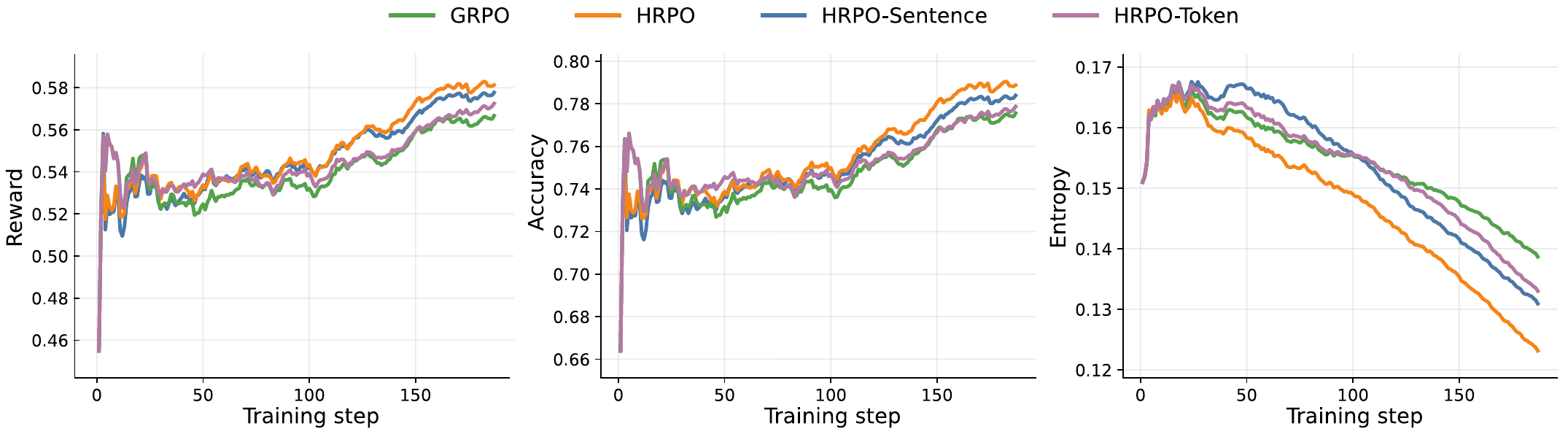}
    \caption{Training dynamics of GRPO and HRPO variants.
From left to right, we show reward, answer accuracy, and policy entropy across training steps.
}

    \label{fig:Acc}
\end{figure*}

\begin{table*}[t]
\centering
\scriptsize
\setlength{\tabcolsep}{3.0pt}
\renewcommand{\arraystretch}{0.92}
\newcommand{\rewpos}[1]{{\normalfont\tiny\textcolor[HTML]{D00000}{(#1\%)}}}
\newcommand{\rewneg}[1]{{\normalfont\tiny\textcolor[HTML]{087F23}{(#1\%)}}}
\resizebox{\textwidth}{!}{%
\begin{tabular}{llccr@{\hspace{0.25em}}lccr@{\hspace{0.25em}}l}
\toprule
Qwen3-1.7B
&
& \multicolumn{4}{c}{GRPO}
& \multicolumn{4}{c}{HRPO} \\
\cmidrule(lr){3-6}
\cmidrule(lr){7-10}
Benchmark & Metric
& Acc reward
& Acc + format
& \multicolumn{2}{c}{Acc + format + length}
& Acc reward
& Acc + format
& \multicolumn{2}{c}{Acc + format + length} \\
\midrule
\multirow{4}{*}{GSM-symbolic P2}
& Acc$\uparrow$
& 0.726 & \textbf{0.762} & 0.756 & \rewneg{-0.8}
& 0.726 & 0.778 & \textbf{0.788} & \rewpos{+1.3} \\
& Streaming Len.$\downarrow$
& \textbf{231.966} & 287.526 & 235.784 & \rewpos{-18.0}
& 231.966 & 275.562 & \textbf{210.046} & \rewpos{-23.8} \\
& Deep Len.$\downarrow$
& 165.942 & 191.756 & \textbf{148.704} & \rewpos{-22.5}
& 165.942 & 160.990 & \textbf{160.210} & \rewpos{-0.5} \\
& Total Len.$\downarrow$
& 397.908 & 479.282 & \textbf{384.488} & \rewpos{-19.8}
& 397.908 & 436.552 & \textbf{370.256} & \rewpos{-15.2} \\
\midrule
\multirow{4}{*}{GSM-symbolic P1}
& Acc$\uparrow$
& 0.850 & \textbf{0.875} & 0.867 & \rewneg{-0.9}
& 0.867 & 0.869 & \textbf{0.871} & \rewpos{+0.2} \\
& Streaming Len.$\downarrow$
& \textbf{168.723} & 206.503 & 172.994 & \rewpos{-16.2}
& 168.723 & 197.902 & \textbf{160.032} & \rewpos{-19.1} \\
& Deep Len.$\downarrow$
& 137.984 & 161.667 & \textbf{133.798} & \rewpos{-17.2}
& 137.984 & 137.218 & \textbf{131.972} & \rewpos{-3.8} \\
& Total Len.$\downarrow$
& \textbf{306.707} & 368.170 & 306.792 & \rewpos{-16.7}
& 306.707 & 335.120 & \textbf{292.004} & \rewpos{-12.9} \\
\bottomrule
\end{tabular}
}
\caption{Analysis of adaptive streaming rewards. We compare GRPO and HRPO under progressively structured rewards, from accuracy-only to format- and length-aware designs. Percentages in \textit{Acc + format + length} are relative to \textit{Acc + format}, with higher accuracy and lower lengths treated as improvements.}
\label{tab:adaptive_rewards}
\end{table*}
Table~\ref{tab:main_results} shows that AdaSR-HRPO improves streaming reasoning by better allocating computation rather than simply generating longer reasoning traces. Compared with the SFT-based StreamingThinker baseline, HRPO consistently improves accuracy on both Qwen3-1.7B and Qwen3-4B, with particularly large gains on harder benchmarks such as GSM-symbolic P2 and MetaMathQA. On Qwen3-1.7B, HRPO improves accuracy by $+22.7\%$ on GSM-symbolic P2 and $+20.1\%$ on MetaMathQA, suggesting that supervised streaming trajectories alone are insufficient for robust reasoning under partial observations.

Compared with GRPO, HRPO achieves a better accuracy-efficiency frontier: across all eight settings, it improves accuracy while reducing total generation length. This indicates that hierarchical advantage assignment alleviates the tendency of sequence-level optimization to reward globally successful but locally inefficient traces. By separating credit across streaming and deep stages, HRPO encourages useful intermediate reasoning while suppressing redundant computation.

The results show a clear model-scale effect. On Qwen3-1.7B, HRPO uses slightly more total tokens than SFT but achieves much higher accuracy, indicating that smaller models benefit from additional adaptive computation. On Qwen3-4B, HRPO improves both accuracy and total length across all benchmarks, reducing total length by $6.9\%$--$13.4\%$ while further increasing accuracy. This suggests that stronger base models can turn adaptive streaming reasoning into a Pareto improvement.

AdaSR's advantage is also better understood from a latency-aware perspective rather than raw total length alone. Unlike read-then-think methods, which defer all reasoning until the full input is observed, AdaSR shifts part of the computation into the streaming stage and keeps final deliberation compact. For example, on GSM-symbolic P2 with Qwen3-1.7B, HRPO reduces deep-stage reasoning from $1866.474$ to $160.210$ tokens while improving accuracy from $0.424$ to $0.788$, showing that AdaSR learns to reason while reading instead of postponing computation to the final response.

\paragraph{Influence of Hierarchical Advantage Assignment}

Table~\ref{tab:hierarchical_advantage} examines how the granularity of advantage assignment affects streaming reasoning. GRPO applies a single sequence-level advantage to all generated tokens, while HRPO separates credit assignment into streaming and deep reasoning stages. We further compare two finer-grained variants: HRPO-sentence, which assigns streaming advantages at the sentence level, and HRPO-token, which applies token-level assignment only to the boundary token associated with the format reward. Other reward components are not token-local by nature, and therefore follow the same assignment scheme as HRPO-sentence.

The results show that stage-level HRPO achieves the best overall accuracy-efficiency trade-off. Compared with GRPO, HRPO improves accuracy on both GSM-symbolic P2 and P1 while reducing total generation length, indicating that separating streaming and deep-stage credit can effectively mitigate the coarse credit assignment of standard GRPO. In contrast, finer-grained assignment does not consistently improve performance. HRPO-sentence reduces deep reasoning length but lowers accuracy, suggesting that sentence-level credit may over-localize reasoning signals and weaken cross-sentence reasoning coherence.

HRPO-token achieves the shortest total length, but its accuracy is lower than HRPO. This indicates that token-level assignment is useful for precise format control, such as boundary-token supervision, but is insufficient for assigning semantic reasoning credit. Overall, the results suggest that the most effective granularity is not the finest one, but the one that matches the natural structure of streaming reasoning: stage-level assignment between streaming and deep deliberation.

Figure~\ref{fig:Acc} shows training dynamics that are consistent with the final evaluation.
The left panel reports the total training reward, which combines accuracy, format, and length rewards, while the middle panel tracks answer accuracy.
Stage-level HRPO achieves the strongest reward curve and the best final accuracy among the compared variants, indicating that separating credit between streaming and deep reasoning provides a more effective optimization signal.
In contrast, the finer-grained variants do not consistently improve the learning curves, suggesting that overly localized assignment may fragment the semantic credit needed for reasoning.
The entropy curves show an initial exploration phase followed by a steady decrease, indicating that the policies gradually become more confident as training progresses.

\paragraph{Analysis of Adaptive Streaming Rewards}
\begin{table*}[t]
\centering
\small
\setlength{\tabcolsep}{5.0pt}
\renewcommand{\arraystretch}{0.95}
\resizebox{\textwidth}{!}{%
\begin{tabular}{lcccccccc}
\toprule
Qwen3-4B
& \multicolumn{4}{c}{GSM-Infinite}
& \multicolumn{4}{c}{LogicNLI} \\
\cmidrule(lr){2-5}
\cmidrule(lr){6-9}
Method
& Acc$\uparrow$ & Streaming Len.$\downarrow$ & Deep Len.$\downarrow$ & Total Len.$\downarrow$
& Acc$\uparrow$ & Streaming Len.$\downarrow$ & Deep Len.$\downarrow$ & Total Len.$\downarrow$ \\
\midrule
SFT-based
& 0.479 & \textbf{157.87} & \textbf{115.54} & 273.408
& 0.474 & \textbf{432.131} & 132.102 & \textbf{564.248} \\
GRPO
& 0.509 & 169.11 & 116.27 & 285.38
& 0.445 & 473.971 & \textbf{121.277} & 595.263 \\
HRPO
& \textbf{0.546} & 134.03 & 131.62 & \textbf{265.65}
& \textbf{0.489} & 462.599 & 129.380 & 591.985 \\
\bottomrule
\end{tabular}
}
\caption{Out-of-distribution on GSM-Infinite and out-of-task on LogicNLI performance with Qwen3-4B. We report final answer accuracy, streaming reasoning length, deep reasoning length, and total generated length. Higher accuracy is better, while lower length indicates more efficient reasoning.}
\label{tab:ood_gsm_infinite_logicnli}
\end{table*}
\begin{table*}[t]
\centering
\scriptsize
\setlength{\tabcolsep}{4.0pt}
\renewcommand{\arraystretch}{0.92}
\newcommand{\latsp}[1]{{\normalfont\tiny\textcolor[HTML]{D00000}{(#1$\times$)}}}
\resizebox{\textwidth}{!}{%
\begin{tabular}{lcccr@{\hspace{0.25em}}lcr@{\hspace{0.25em}}lcr@{\hspace{0.25em}}l}
\toprule
Backend
& \multicolumn{2}{c}{Read-then-think}
& \multicolumn{3}{c}{StreamingThinker}
& \multicolumn{3}{c}{AdaSR-GRPO}
& \multicolumn{3}{c}{AdaSR-HRPO} \\
\cmidrule(lr){2-3}
\cmidrule(lr){4-6}
\cmidrule(lr){7-9}
\cmidrule(lr){10-12}
& Throughput$\uparrow$
& Latency$\downarrow$
& Throughput$\uparrow$
& \multicolumn{2}{c}{Latency$\downarrow$}
& Throughput$\uparrow$
& \multicolumn{2}{c}{Latency$\downarrow$}
& Throughput$\uparrow$
& \multicolumn{2}{c}{Latency$\downarrow$} \\
\midrule
Transformers
& 33.170 & 56.250
& 31.810 & \textbf{6.080} & \latsp{9.25}
& 31.450 & 6.710 & \latsp{8.38}
& 32.570 & 6.520 & \latsp{8.63} \\
vLLM
& 140.600 & 13.270
& 134.509 & \textbf{1.403} & \latsp{9.46}
& 136.319 & 1.595 & \latsp{8.32}
& 136.945 & 1.505 & \latsp{8.82} \\
\bottomrule
\end{tabular}
}
\caption{The throughput (token/s) and reasoning latency (s) under different inference backends. We compare read-then-think, StreamingThinker, AdaSR-GRPO, and AdaSR-HRPO using Transformers and vLLM backends. Throughput is higher-is-better, while latency is lower-is-better. Speedups in parentheses for latency denote reductions relative to read-then-think under the same backend.}
\label{tab:streaming_stage_latency}
\end{table*}

Table~\ref{tab:adaptive_rewards} examines how reward components affect adaptive streaming reasoning, where the model must decide when to reason or skip during the input stream. With accuracy reward alone, the model receives only a final outcome signal and lacks guidance on streaming-format compliance. Adding the format reward improves accuracy under both GRPO and HRPO, suggesting that parseable read-think trajectories stabilize training, though with longer generations.

The length-aware reward encourages more efficient computation allocation across streaming and deep stages. Under GRPO, it reduces generation length but slightly hurts accuracy on GSM-symbolic P2 and P1, indicating that flat sequence-level optimization applies length pressure too coarsely. In contrast, HRPO benefits more consistently: compared with \textit{Acc + format}, adding the length reward improves accuracy while reducing total length by $15.2\%$ on P2 and $12.9\%$ on P1, mainly by suppressing redundant streaming-stage thoughts while preserving deep-stage deliberation.

These results show that adaptive rewards and hierarchical advantage assignment are complementary: format rewards make streaming trajectories learnable, length-aware rewards make them efficient, and HRPO assigns these signals to the appropriate reasoning stages.

\paragraph{Performance on Out-of-Domain Cases}

Table~\ref{tab:ood_gsm_infinite_logicnli} evaluates out-of-distribution generalization on GSM-Infinite and out-of-task generalization on LogicNLI. On GSM-Infinite, both GRPO and HRPO improve accuracy over the SFT-based baseline, indicating that RL optimization transfers to longer mathematical inputs. HRPO achieves the best accuracy, increasing performance from $0.479$ to $0.546$, while also producing the shortest streaming length and total length among all methods. On LogicNLI, HRPO again obtains the highest accuracy, improving over both SFT-based and GRPO models. Although SFT-based decoding remains shorter in total length on LogicNLI, HRPO reduces the total length compared with GRPO while achieving better accuracy, suggesting a stronger performance trade-off under out-of-task generalization.
\paragraph{Analysis of Streaming Latency}
Table~\ref{tab:streaming_stage_latency} reports the practical latency under Transformers and vLLM backends on the GSM-Symbolic dataset.\footnote{Following the StreamingThinker~\cite{tong2025streamingthinker}, we set the input streaming speed to 150 words per minute.}
Compared with read-then-think, streaming reasoning greatly reduces exposed latency, since intermediate reasoning can be overlapped with the arrival of input segments and only the final deep reasoning remains on the critical path. As a result, all streaming methods achieve over $8\times$ latency reduction under both backends. Although AdaSR adaptively reallocates computation between streaming and deep stages, it introduces only minor latency fluctuations compared with StreamingThinker, especially under vLLM. With the vLLM backend, streaming inference is further accelerated, achieving a $4.3\times$ speedup in AdaSR-HRPO latency, which demonstrates that AdaSR can be efficiently deployed in realistic serving scenarios.

\section{Conclusion}

In this paper, we presented AdaSR, an adaptive framework for streaming reasoning that enables LLMs to decide when to think, when to skip, and how to allocate computation between intermediate streaming reasoning and final deep deliberation as inputs progressively unfold. To optimize this structured reasoning process, we introduced Hierarchical Relative Policy Optimization, which decomposes advantage assignment across streaming, deep-reasoning, and global levels, thereby mitigating the coarse credit assignment inherent in sequence-level optimization. Experiments demonstrate that AdaSR achieves a better trade-off among reasoning accuracy, computational efficiency, and streaming latency. These results suggest that streaming reasoning is not merely a variant of conventional long-form reasoning, but a distinct optimization problem that requires temporal credit assignment and adaptive computation allocation. We hope this work provides a step toward more flexible, efficient, and responsive reasoning models for real-time and dynamically evolving inputs.

\clearpage
\section*{Limitations}
AdaSR is an initial step toward RL-based adaptive streaming reasoning. We focus on text streams with verifiable answers, which allows us to isolate the temporal credit-assignment problem and evaluate HRPO cleanly. Extending this framework to continuous audio, video, and more open-ended interactive streams is a promising next direction, likely requiring richer reward signals, modality-specific rollout engines, and adaptive scheduling of the hierarchy and reward-shaping coefficients.
\bibliography{ref}

@article{tong2025streamingthinker,
  title={StreamingThinker: Large Language Models Can Think While Reading},
  author={Tong, Junlong and Fan, Yingqi and Zhao, Anhao and Ma, Yunpu and Shen, Xiaoyu},
  journal={arXiv preprint arXiv:2510.17238},
  year={2025}
}

@article{ma2018stacl,
  title={STACL: Simultaneous Translation with Implicit Anticipation and Controllable Latency using Prefix-to-Prefix Framework},
  author={Ma, Mingbo and Huang, Liang and Xiong, Hao and Zheng, Renjie and Liu, Kaibo and Zheng, Baigong and Zhang, Chuanqiang and He, Zhongjun and Liu, Hairong and Li, Xing and others},
  journal={arXiv preprint arXiv:1810.08398},
  year={2018}
}

@article{raffel2024simultaneous,
  title={Simultaneous Masking, Not Prompting Optimization: A Paradigm Shift in Fine-Tuning LLMs for Simultaneous Translation},
  author={Raffel, Matthew and Agostinelli, Victor and Chen, Lizhong},
  journal={arXiv preprint arXiv:2405.10443},
  year={2024}
}

@article{guo2024decoder,
  title={Decoder-Only Streaming Transformer for Simultaneous Translation},
  author={Guo, Shoutao and Zhang, Shaolei and Feng, Yang},
  journal={arXiv preprint arXiv:2406.03878},
  year={2024}
}

@article{tong2025llm,
  title={LLM as Effective Streaming Processor: Bridging Streaming-Batch Mismatches with Group Position Encoding},
  author={Tong, Junlong and Fu, Jinlan and Lin, Zixuan and Fan, Yingqi and Zhao, Anhao and Su, Hui and Shen, Xiaoyu},
  journal={arXiv preprint arXiv:2505.16983},
  year={2025}
}

@inproceedings{chen2024videollm,
  title={VideoLLM-Online: Online Video Large Language Model for Streaming Video},
  author={Chen, Joya and Lv, Zhaoyang and Wu, Shiwei and Lin, Kevin Qinghong and Song, Chenan and Gao, Difei and Liu, Jia-Wei and Gao, Ziteng and Mao, Dongxing and Shou, Mike Zheng},
  booktitle={Proceedings of the IEEE/CVF Conference on Computer Vision and Pattern Recognition},
  pages={18407--18418},
  year={2024}
}

@inproceedings{chen2025livecc,
  title={LiveCC: Learning Video LLM with Streaming Speech Transcription at Scale},
  author={Chen, Joya and Zeng, Ziyun and Lin, Yiqi and Li, Wei and Ma, Zejun and Shou, Mike Zheng},
  booktitle={Proceedings of the Computer Vision and Pattern Recognition Conference},
  pages={29083--29095},
  year={2025}
}

@article{xie2024mini,
  title={Mini-Omni: Language Models Can Hear, Talk While Thinking in Streaming},
  author={Xie, Zhifei and Wu, Changqiao},
  journal={arXiv preprint arXiv:2408.16725},
  year={2024}
}

@article{lin2026speak,
  title={Speak While Watching: Unleashing TRUE Real-Time Video Understanding Capability of Multimodal Large Language Models},
  author={Lin, Junyan and Tong, Junlong and Wu, Hao and Zhang, Jialiang and Liu, Jinming and Jin, Xin and Shen, Xiaoyu},
  journal={arXiv preprint arXiv:2601.06843},
  year={2026}
}

@article{xie2025interleaved,
  title={Interleaved Reasoning for Large Language Models via Reinforcement Learning},
  author={Xie, Roy and Qiu, David and Gopinath, Deepak and Lin, Dong and Sun, Yanchao and Wang, Chong and Potdar, Saloni and Dhingra, Bhuwan},
  journal={arXiv preprint arXiv:2505.19640},
  year={2025}
}

@article{chiang2025stitch,
  title={STITCH: Simultaneous Thinking and Talking with Chunked Reasoning for Spoken Language Models},
  author={Chiang, Cheng-Han and Wang, Xiaofei and Li, Linjie and Lin, Chung-Ching and Lin, Kevin and Liu, Shujie and Wang, Zhendong and Yang, Zhengyuan and Lee, Hung-yi and Wang, Lijuan},
  journal={arXiv preprint arXiv:2507.15375},
  year={2025}
}

@article{xie2025mini,
  title={Mini-Omni-Reasoner: Token-Level Thinking-in-Speaking in Large Speech Models},
  author={Xie, Zhifei and Ma, Ziyang and Liu, Zihang and Pang, Kaiyu and Li, Hongyu and Zhang, Jialin and Liao, Yue and Ye, Deheng and Miao, Chunyan and Yan, Shuicheng},
  journal={arXiv preprint arXiv:2508.15827},
  year={2025}
}

@article{liu2024streamchat,
  title={StreamChat: Chatting with Streaming Video},
  author={Liu, Jihao and Yu, Zhiding and Lan, Shiyi and Wang, Shihao and Fang, Rongyao and Kautz, Jan and Li, Hongsheng and Alvare, Jose M},
  journal={arXiv preprint arXiv:2412.08646},
  year={2024}
}

@article{zhang2026tays,
  title={Think-as-You-See: Streaming Chain-of-Thought Reasoning for Large Vision-Language Models},
  author={Zhang, Jialiang and Tong, Junlong and Lin, Junyan and Wu, Hao and Sun, Yirong and Ma, Yunpu and Shen, Xiaoyu},
  journal={arXiv preprint arXiv:2603.02872},
  year={2026}
}

@article{gim2024asynchronous,
  title={Asynchronous LLM Function Calling},
  author={Gim, In and Lee, Seung-seob and Zhong, Lin},
  journal={arXiv preprint arXiv:2412.07017},
  year={2024}
}

@article{arora2025stream,
  title={Stream RAG: Instant and Accurate Spoken Dialogue Systems with Streaming Tool Usage},
  author={Arora, Siddhant and Khan, Haidar and Sun, Kai and Dong, Xin Luna and Choudhary, Sajal and Moon, Seungwhan and Zhang, Xinyuan and Sagar, Adithya and Appini, Surya Teja and Patnaik, Kaushik and others},
  journal={arXiv preprint arXiv:2510.02044},
  year={2025}
}

@article{zhang2025avila,
  title={AViLA: Asynchronous Vision-Language Agent for Streaming Multimodal Data Interaction},
  author={Zhang, Gengyuan and Hannan, Tanveer and Kleiner, Hermine and Aydemir, Beste and Xie, Xinyu and Lan, Jian and Seidl, Thomas and Tresp, Volker and Gu, Jindong},
  journal={arXiv preprint arXiv:2506.18472},
  year={2025}
}

@article{yang2025streamagent,
  title={StreamAgent: Towards Anticipatory Agents for Streaming Video Understanding},
  author={Yang, Haolin and Tang, Feilong and Zhao, Lingxiao and An, Xiang and Hu, Ming and Li, Huifa and Zhuang, Xinlin and Lu, Yifan and Zhang, Xiaofeng and Swikir, Abdalla and others},
  journal={arXiv preprint arXiv:2508.01875},
  year={2025}
}

@inproceedings{arivazhagan2019monotonic,
  title={Monotonic Infinite Lookback Attention for Simultaneous Machine Translation},
  author={Arivazhagan, Naveen and Cherry, Colin and Macherey, Wolfgang and Chiu, Chung-Cheng and Yavuz, Semih and Pang, Ruoming and Li, Wei and Raffel, Colin},
  booktitle={Proceedings of the 57th Annual Meeting of the Association for Computational Linguistics},
  pages={1313--1323},
  year={2019}
}

@inproceedings{elbayad2020efficient,
  title={Efficient Wait-k Models for Simultaneous Machine Translation},
  author={Elbayad, Maha and Besacier, Laurent and Verbeek, Jakob},
  booktitle={Proceedings of Interspeech 2020},
  year={2020}
}

@inproceedings{agostinelli2024simul,
  title={Simul-LLM: A Framework for Exploring High-Quality Simultaneous Translation with Large Language Models},
  author={Agostinelli, Victor and Wild, Max and Raffel, Matthew and Fuad, Kazi Ahmed Asif and Chen, Lizhong},
  booktitle={Proceedings of the 62nd Annual Meeting of the Association for Computational Linguistics (Volume 1: Long Papers)},
  year={2024}
}

@article{guo2024agent,
  title={Agent-SiMT: Agent-assisted Simultaneous Machine Translation with Large Language Models},
  author={Guo, Shoutao and Zhang, Shaolei and Ma, Zhengrui and Zhang, Min and Feng, Yang},
  journal={arXiv preprint arXiv:2406.06910},
  year={2024}
}

@article{ouyang2024anticipating,
  title={Anticipating Future with Large Language Model for Simultaneous Machine Translation},
  author={Ouyang, Siqi and Hrinchuk, Oleksii and Chen, Zhehuai and Lavrukhin, Vitaly and Balam, Jagadeesh and Li, Lei and Ginsburg, Boris},
  journal={arXiv preprint arXiv:2410.22499},
  year={2024}
}

@article{lin2024streamingbench,
  title={StreamingBench: Assessing the Gap for MLLMs to Achieve Streaming Video Understanding},
  author={Lin, Junming and Fang, Zheng and Chen, Chi and Wan, Zihao and Luo, Fuwen and Li, Peng and Liu, Yang and Sun, Maosong},
  journal={arXiv preprint arXiv:2411.03628},
  year={2024}
}

@article{wang2026think,
  title={Think While Watching: Online Streaming Segment-Level Memory for Multi-Turn Video Reasoning in Multimodal Large Language Models},
  author={Wang, Lu and Jin, Zhuoran and Hao, Yupu and Chen, Yubo and Liu, Kang and Ao, Yulong and Zhao, Jun},
  journal={arXiv preprint arXiv:2603.11896},
  year={2026}
}

@article{defossez2024moshi,
  title={Moshi: A Speech-Text Foundation Model for Real-Time Dialogue},
  author={D{\'e}fossez, Alexandre and Mazar{\'e}, Laurent and Orsini, Manu and Royer, Am{\'e}lie and P{\'e}rez, Patrick and J{\'e}gou, Herv{\'e} and Grave, Edouard and Zeghidour, Neil},
  journal={arXiv preprint arXiv:2410.00037},
  year={2024}
}

@article{fang2024llamaomni,
  title={LLaMA-Omni: Seamless Speech Interaction with Large Language Models},
  author={Fang, Qingkai and Guo, Shoutao and Zhou, Yan and Ma, Zhengrui and Zhang, Shaolei and Feng, Yang},
  journal={arXiv preprint arXiv:2409.06666},
  year={2024}
}

@article{zeng2024glm4voice,
  title={GLM-4-Voice: Towards Intelligent and Human-Like End-to-End Spoken Chatbot},
  author={Zeng, Aohan and Du, Zhengxiao and Liu, Mingdao and Wang, Kedong and Jiang, Shengmin and Zhao, Lei and Dong, Yuxiao and Tang, Jie},
  journal={arXiv preprint arXiv:2412.02612},
  year={2024}
}

@inproceedings{zhang2024omniflatten,
  title={OmniFlatten: An End-to-End GPT Model for Seamless Voice Conversation},
  author={Zhang, Qinglin and Cheng, Luyao and Deng, Chong and Chen, Qian and Wang, Wen and Zheng, Siqi and Liu, Jiaqing and Yu, Hai and Tan, Chaohong and Du, Zhihao and Zhang, Shiliang},
  booktitle={Proceedings of the 63rd Annual Meeting of the Association for Computational Linguistics (Volume 1: Long Papers)},
  year={2025}
}

@article{yu2024salmonnomni,
  title={SALMONN-omni: A Codec-free LLM for Full-duplex Speech Understanding and Generation},
  author={Yu, Wenyi and Wang, Siyin and Yang, Xiaoyu and Chen, Xianzhao and Tian, Xiaohai and Zhang, Jun and Sun, Guangzhi and Lu, Lu and Wang, Yuxuan and Zhang, Chao},
  journal={arXiv preprint arXiv:2411.18138},
  year={2024}
}

@article{chiang2025shanks,
  title={SHANKS: Simultaneous Hearing and Thinking for Spoken Language Models},
  author={Chiang, Cheng-Han and Wang, Xiaofei and Li, Linjie and Lin, Chung-Ching and Lin, Kevin and Liu, Shujie and Wang, Zhendong and Yang, Zhengyuan and Lee, Hung-yi and Wang, Lijuan},
  journal={arXiv preprint arXiv:2510.06917},
  year={2025}
}

@article{liu2026ddtsr,
  title={Discourse-Aware Dual-Track Streaming Response for Low-Latency Spoken Dialogue Systems},
  author={Liu, Siyuan and Xu, Jiahui and Jiang, Feng and Wang, Kuang and Zhao, Zefeng and Huang, Chu-Ren and Gu, Jinghang and Yin, Changqing and Li, Haizhou},
  journal={arXiv preprint arXiv:2602.23266},
  year={2026}
}

@inproceedings{yao2023react,
  title={ReAct: Synergizing Reasoning and Acting in Language Models},
  author={Yao, Shunyu and Zhao, Jeffrey and Yu, Dian and Du, Nan and Shafran, Izhak and Narasimhan, Karthik and Cao, Yuan},
  booktitle={Proceedings of the 11th International Conference on Learning Representations},
  year={2023}
}

@inproceedings{hu2023thought,
  title={Thought Cloning: Learning to Think while Acting by Imitating Human Thinking},
  author={Hu, Shengran and Clune, Jeff},
  booktitle={Advances in Neural Information Processing Systems},
  year={2023}
}

@article{shao2024deepseekmath,
  title={DeepSeekMath: Pushing the Limits of Mathematical Reasoning in Open Language Models},
  author={Shao, Zhihong and Wang, Peiyi and Zhu, Qihao and Xu, Runxin and Song, Junxiao and Li, Xiao and Zhang, Mingchuan and Zhang, Y.K. and Li, Yu and Wu, Y. and Guo, Daya},
  journal={arXiv preprint arXiv:2402.03300},
  year={2024}
}

@article{deepseek2025r1,
  title={DeepSeek-R1: Incentivizing Reasoning Capability in LLMs via Reinforcement Learning},
  author={{DeepSeek-AI}},
  journal={arXiv preprint arXiv:2501.12948},
  year={2025}
}

@article{wei2022chain,
  title={Chain-of-Thought Prompting Elicits Reasoning in Large Language Models},
  author={Wei, Jason and Wang, Xuezhi and Schuurmans, Dale and Bosma, Maarten and Ichter, Brian and Xia, Fei and Chi, Ed and Le, Quoc V and Zhou, Denny},
  journal={Advances in Neural Information Processing Systems},
  volume={35},
  pages={24824--24837},
  year={2022}
}

@article{schulman2017proximal,
  title={Proximal Policy Optimization Algorithms},
  author={Schulman, John and Wolski, Filip and Dhariwal, Prafulla and Radford, Alec and Klimov, Oleg},
  journal={arXiv preprint arXiv:1707.06347},
  year={2017}
}

@article{ouyang2022training,
  title={Training Language Models to Follow Instructions with Human Feedback},
  author={Ouyang, Long and Wu, Jeffrey and Jiang, Xu and Almeida, Diogo and Wainwright, Carroll and Mishkin, Pamela and Zhang, Chong and Agarwal, Sandhini and Slama, Katarina and Ray, Alex and others},
  journal={Advances in Neural Information Processing Systems},
  volume={35},
  pages={27730--27744},
  year={2022}
}

@article{kwon2023efficient,
  title={Efficient Memory Management for Large Language Model Serving with PagedAttention},
  author={Kwon, Woosuk and Li, Zhuohan and Zhuang, Siyuan and Sheng, Ying and Zheng, Lianmin and Yu, Cody Hao and Gonzalez, Joseph and Zhang, Hao and Stoica, Ion},
  journal={Proceedings of the 29th Symposium on Operating Systems Principles},
  pages={611--626},
  year={2023}
}

@article{openai2024o1,
  title={Learning to Reason with LLMs},
  author={{OpenAI}},
  journal={OpenAI Blog},
  year={2024}
}

@article{ahmadian2024back,
  title={Back to Basics: Revisiting REINFORCE Style Optimization for Learning from Human Feedback in LLMs},
  author={Ahmadian, Arash and Cremer, Chris and Gall{\'e}, Matthias and Fadaee, Marzieh and Kreutzer, Julia and Pietquin, Olivier and {\"U}st{\"u}n, Ahmet and Hooker, Sara},
  journal={arXiv preprint arXiv:2402.14740},
  year={2024}
}

@article{su2024roformer,
  title={RoFormer: Enhanced Transformer with Rotary Position Embedding},
  author={Su, Jianlin and Ahmed, Murtadha and Lu, Yu and Pan, Shengfeng and Bo, Wen and Liu, Yunfeng},
  journal={Neurocomputing},
  volume={568},
  pages={127063},
  year={2024}
}

@article{snell2024scaling,
  title={Scaling LLM Test-Time Compute Optimally Can Be More Effective Than Scaling Model Parameters},
  author={Snell, Charlie and Lee, Jaehoon and Xu, Kelvin and Kumar, Aviral},
  journal={arXiv preprint arXiv:2408.03314},
  year={2024}
}

@article{sheng2024hybridflow,
  title={HybridFlow: A Flexible and Efficient RLHF Framework},
  author={Sheng, Guangming and Zhang, Chi and Ye, Zilingfeng and Wu, Xibin and Zhang, Wang and Zhang, Ru and Peng, Yanghua and Lin, Haibin and Wu, Chuan},
  journal={arXiv preprint arXiv:2409.19256},
  year={2024}
}

@article{liu2026gdpo,
  title={GDPO: Group reward-Decoupled Normalization Policy Optimization for Multi-reward RL Optimization},
  author={Liu, Shih-Yang and Dong, Xin and Lu, Ximing and Diao, Shizhe and Belcak, Peter and Liu, Mingjie and Chen, Min-Hung and Yin, Hongxu and Wang, Yu-Chiang Frank and Cheng, Kwang-Ting and Choi, Yejin and Kautz, Jan and Molchanov, Pavlo},
  journal={arXiv preprint arXiv:2601.05242},
  year={2026}
}

@article{cobbe2021training,
  title={Training Verifiers to Solve Math Word Problems},
  author={Cobbe, Karl and Kosaraju, Vineet and Bavarian, Mohammad and Chen, Mark and Jun, Heewoo and Kaiser, Lukasz and Plappert, Matthias and Tworek, Jerry and Hilton, Jacob and Nakano, Reiichiro and Hesse, Christopher and Schulman, John},
  journal={arXiv preprint arXiv:2110.14168},
  year={2021}
}

@article{graves2016adaptive,
  title={Adaptive Computation Time for Recurrent Neural Networks},
  author={Graves, Alex},
  journal={arXiv preprint arXiv:1603.08983},
  year={2016}
}

@inproceedings{gu2017learning,
  title={Learning to Translate in Real-time with Neural Machine Translation},
  author={Gu, Jiatao and Neubig, Graham and Cho, Kyunghyun and Li, Victor O.K.},
  booktitle={Proceedings of the 15th Conference of the European Chapter of the Association for Computational Linguistics: Volume 1, Long Papers},
  pages={1053--1062},
  year={2017},
  address={Valencia, Spain},
  publisher={Association for Computational Linguistics}
}

@article{ziegler2019fine,
  title={Fine-Tuning Language Models from Human Preferences},
  author={Ziegler, Daniel M. and Stiennon, Nisan and Wu, Jeffrey and Brown, Tom B. and Radford, Alec and Amodei, Dario and Christiano, Paul and Irving, Geoffrey},
  journal={arXiv preprint arXiv:1909.08593},
  year={2019}
}

@article{muennighoff2025s1,
  title={s1: Simple Test-Time Scaling},
  author={Muennighoff, Niklas and Yang, Zitong and Shi, Weijia and Li, Xiang Lisa and Fei-Fei, Li and Hajishirzi, Hannaneh and Zettlemoyer, Luke and Liang, Percy and Cand{\`e}s, Emmanuel and Hashimoto, Tatsunori},
  journal={arXiv preprint arXiv:2501.19393},
  year={2025}
}

@article{liu2025understanding,
  title={Understanding R1-Zero-Like Training: A Critical Perspective},
  author={Liu, Zichen and Chen, Changyu and Li, Wenjun and Qi, Penghui and Pang, Tianyu and Du, Chao and Lee, Wee Sun and Lin, Min},
  journal={arXiv preprint arXiv:2503.20783},
  year={2025}
}

@article{yu2025dapo,
  title={DAPO: An Open-Source LLM Reinforcement Learning System at Scale},
  author={Yu, Qiying and Zhang, Zheng and Zhu, Ruofei and Yuan, Yufeng and Zuo, Xiaochen and Yue, Yu and Dai, Weinan and Fan, Tiantian and Liu, Gaohong and Liu, Lingjun and others},
  journal={arXiv preprint arXiv:2503.14476},
  year={2025}
}

@article{wang2025perception,
  title={Perception-Aware Policy Optimization for Multimodal Reasoning},
  author={Wang, Zhenhailong and Guo, Xuehang and Stoica, Sofia and Xu, Haiyang and Wang, Hongru and Ha, Hyeonjeong and Chen, Xiusi and Chen, Yangyi and Yan, Ming and Huang, Fei and others},
  journal={arXiv preprint arXiv:2507.06448},
  year={2025}
}

@article{mirzadeh2024gsmsymbolic,
  title={GSM-Symbolic: Understanding the Limitations of Mathematical Reasoning in Large Language Models},
  author={Mirzadeh, Iman and Alizadeh, Keivan and Shahrokhi, Hooman and Tuzel, Oncel and Bengio, Samy and Farajtabar, Mehrdad},
  journal={arXiv preprint arXiv:2410.05229},
  year={2024}
}

@article{yu2023metamath,
  title={MetaMath: Bootstrap Your Own Mathematical Questions for Large Language Models},
  author={Yu, Longhui and Jiang, Weisen and Shi, Han and Yu, Jincheng and Liu, Zhengying and Zhang, Yu and Kwok, James T. and Li, Zhenguo and Weller, Adrian and Liu, Weiyang},
  journal={arXiv preprint arXiv:2309.12284},
  year={2023}
}

@inproceedings{jin2019pubmedqa,
  title={PubMedQA: A Dataset for Biomedical Research Question Answering},
  author={Jin, Qiao and Dhingra, Bhuwan and Liu, Zhengping and Cohen, William W. and Lu, Xinghua},
  booktitle={Proceedings of the 2019 Conference on Empirical Methods in Natural Language Processing and the 9th International Joint Conference on Natural Language Processing},
  pages={2567--2577},
  year={2019}
}

@article{zhou2025gsminfinite,
  title={GSM-Infinite: How Do Your LLMs Behave over Infinitely Increasing Context Length and Reasoning Complexity?},
  author={Zhou, Yang and Liu, Hongyi and Chen, Zhuoming and Tian, Yuandong and Chen, Beidi},
  journal={arXiv preprint arXiv:2502.05252},
  year={2025}
}

@article{yang2025qwen3,
  title={Qwen3 Technical Report},
  author={Yang, An and Li, Anfeng and Yang, Baosong and Zhang, Beichen and Hui, Binyuan and Zheng, Bo and Yu, Bowen and Gao, Chang and Huang, Chengen and Lv, Chenxu and Zheng, Chujie and Liu, Dayiheng and Zhou, Fan and Huang, Fei and Hu, Feng and Ge, Hao and Wei, Haoran and Lin, Huan and Tang, Jialong and Yang, Jian and Tu, Jianhong and Zhang, Jianwei and others},
  journal={arXiv preprint arXiv:2505.09388},
  year={2025}
}

@inproceedings{tian-etal-2021-diagnosing,
    title = "Diagnosing the First-Order Logical Reasoning Ability Through {L}ogic{NLI}",
    author = "Tian, Jidong  and
      Li, Yitian  and
      Chen, Wenqing  and
      Xiao, Liqiang  and
      He, Hao  and
      Jin, Yaohui",
    year = "2021",
    pages = "3738--3747",
}

@article{tong2026proactivellm,
  title={ProactiveLLM: Learning Active Interaction for Streaming Large Language Models},
  author={Tong, Junlong and Zhang, Yao and Zhao, Anhao and Fan, Yingqi and Ma, Yunpu and Shen, Xiaoyu},
  journal={arXiv preprint arXiv:2606.00523},
  year={2026}
}

@article{tong2026static,
  title={From static inference to dynamic interaction: A survey of streaming large language models},
  author={Tong, Junlong and Wang, Zilong and Ren, YuJie and Yin, Peiran and Wu, Hao and Zhang, Wei and Shen, Xiaoyu},
  journal={arXiv preprint arXiv:2603.04592},
  year={2026}
}

\appendix

\newpage
\newpage

\section{Related Work}

\paragraph{Streaming LLMs}
Streaming LLMs~\cite{tong2026static} aim to move language-model inference from static full-context processing toward dynamic interaction, where input perception, reasoning, output generation, and external actions may overlap. From an application perspective, existing work has explored several forms of real-time interaction. Simultaneous translation studies \emph{read while outputting}, where models learn or impose read/write policies such as wait-$k$ and efficient interaction decision to trade source completeness for latency~\citep{ma2018stacl,arivazhagan2019monotonic,elbayad2020efficient,tong2025llm,tong2026proactivellm}, with recent LLM-based and speech-to-speech systems extending this setting to real-time translation~\citep{agostinelli2024simul,raffel2024simultaneous,guo2024decoder,guo2024agent,ouyang2024anticipating}. Streaming video and multimodal systems study \emph{think/speak while watching}, processing frames incrementally, chatting over evolving video streams, or generating reasoning synchronized with incoming visual evidence~\citep{chen2024videollm,liu2024streamchat,chen2025livecc,lin2026speak,zhang2026tays,wang2026think}. Spoken dialogue systems study \emph{think while hearing} and \emph{speak while thinking}, enabling low-latency or full-duplex speech interaction and, in some cases, hidden reasoning during listening or speaking~\citep{xie2024mini,defossez2024moshi,fang2024llamaomni,zeng2024glm4voice,zhang2024omniflatten,yu2024salmonnomni,chiang2025stitch,xie2025mini,chiang2025shanks,liu2026ddtsr}. Agentic systems further study \emph{think while acting}, where models interleave reasoning with tool use, retrieval, function calls, or anticipatory actions under streaming observations~\citep{yao2023react,hu2023thought,gim2024asynchronous,arora2025stream,zhang2025avila,yang2025streamagent,xie2025interleaved}. These scenarios show that streaming LLMs are motivated not only by faster token emission, but also by the need to coordinate computation with continuously evolving inputs.

Among these scenarios, \emph{think while reading} provides a clean abstraction for streaming reasoning, where the model processes input segments sequentially, reasons under partial observations, and performs final deliberation after the stream ends. StreamingThinker~\citep{tong2025streamingthinker} realizes this paradigm with sentence-level streaming CoT and streaming-specific attention, position encoding, and KV-cache designs, showing how LLMs can reason under streaming visibility constraints. Yet its intermediate reasoning behavior is largely shaped by constructed supervised traces. AdaSR instead studies whether models can learn their own computation policy: when to think, when to skip, and how to allocate computation between online thoughts and final deep reasoning.

\paragraph{Policy Optimization in RLVR}
Reinforcement Learning with Verifiable Rewards (RLVR) has recently become a
central paradigm for improving the reasoning ability of large language models,
where policies are optimized using automatically checkable outcome rewards rather
than learned reward models~\citep{cobbe2021training,shao2024deepseekmath,deepseek2025r1}.
Early RLHF and RLVR systems commonly build upon PPO-style clipped policy
objectives~\citep{schulman2017proximal,ziegler2019fine,ouyang2022training},
while GRPO replaces the critic with group-relative reward normalization,
estimating the baseline from multiple rollouts of the same prompt and thereby
reducing the cost of value modeling~\citep{shao2024deepseekmath}. Recent studies
further refine this paradigm from different perspectives. REINFORCE-style
methods revisit critic-free optimization for language-model
alignment~\citep{ahmadian2024back}. Dr. GRPO analyzes the optimization bias of GRPO and
shows that length-dependent loss aggregation can artificially encourage longer
responses, especially for incorrect trajectories; it therefore modifies the
aggregation and normalization scheme to improve token efficiency while preserving
reasoning performance~\citep{liu2025understanding}. DAPO improves the
scalability and stability of long-CoT RL by introducing decoupled clipping and
dynamic sampling, together with practical techniques such as Clip-Higher,
token-level policy-gradient loss, and overlong reward shaping~\citep{yu2025dapo}.
GDPO extends GRPO to multi-reward RLVR by decoupling the normalization of
different reward components before aggregation, alleviating advantage collapse
caused by directly normalizing summed rewards~\citep{liu2026gdpo}. Beyond
text-only reasoning, PAPO incorporates perception-aware optimization into RLVR by
adding an implicit perception loss and entropy regularization, enabling GRPO- or
DAPO-style training to better support multimodal
reasoning~\citep{wang2025perception}. Overall, these methods improve RLVR along
dimensions such as critic-free optimization, length-bias correction, training
stability, multi-reward composition, and multimodal grounding. However, they
still largely optimize completed trajectories at the sequence or token level,
without explicitly modeling the hierarchical temporal structure of streaming
reasoning. In contrast, our HRPO targets the phase-structured credit assignment
problem, assigning advantages over streaming, deep-reasoning, and global token
ranges to better optimize adaptive reasoning under partial and evolving inputs.

\section{HRPO Analysis}

\paragraph{GRPO Limitation for Streaming Reasoning}
We further formalize why the standard GRPO objective in
Eq.~\ref{eq:grpo} is mismatched with streaming reasoning. For a sampled
streaming trajectory $o_i=[R_1,\ldots,R_T,R]$, let $t_{\mathrm{bnd}}$ denote
the boundary between the online streaming phase and the final deep-thinking
phase, as defined in Eq.~\ref{eq:paradigm}. GRPO first collapses the whole
trajectory into a scalar reward $R_i$ and then assigns the resulting
group-normalized advantage $\hat{A}_i$ to every token:
{
\begin{equation}
\label{eq:grpo_flat_gradient}
\begin{aligned}
&\nabla_{\theta}J_{\mathrm{GRPO}}(\theta)
\propto
\mathbb{E}\!\Bigg[
\frac{1}{G}\sum_{i=1}^{G}
\frac{1}{|o_i|}
\\
&\sum_{t=1}^{|o_i|}
m_{i,t} r_{i,t}\hat{A}_i
\nabla_{\theta}\log
\pi_{\theta}\!\left(
o_{i,t}\mid q,o_{i,<t}
\right)
\Bigg].
\end{aligned}
\end{equation}
}
where $m_{i,t}$ denotes whether the clipped surrogate is active. Thus tokens
before and after $t_{\mathrm{bnd}}$ receive the same learning signal even
though they play different roles: streaming tokens should be judged by the
quality and timeliness of partial-observation reasoning, while deep tokens
should be judged by full-context integration. \textbf{This flat credit assignment creates two failure modes.} If an incorrect or unhelpful streaming segment is rescued by deep deliberation, GRPO still reinforces the streaming tokens
because the final reward is high; conversely, useful streaming thoughts are
penalized whenever the final answer fails for reasons in the deep phase.
Consequently, GRPO cannot tell whether performance changes come from online
reasoning, final deliberation, or their interaction, motivating the
hierarchical advantage decomposition used by HRPO.

\paragraph{Advantage Decomposition and Policy Gradient}
\label{app:hrpo-gradient}
We keep the notation in Eq.~\ref{eq:hrpo}. Starting from the original HRPO
objective, the local and global surrogate terms can be grouped by token time step. Streaming tokens receive the streaming local signal and the global signal; deep tokens receive the deep local signal and the same global signal:

\begin{equation}
\label{eq:hrpo_time_grouped}
\begin{aligned}
&J_{\mathrm{HRPO}}(\theta)
=
\mathbb{E}_{\substack{
(q,a) \sim \mathcal{D} \\
\{o_i\}_{i=1}^{G} \sim \pi_{\theta_{\mathrm{old}}}
}}
\Bigg\{
\frac{1}{G}\sum_{i=1}^{G}
\Bigg[
\\
&\resizebox{\columnwidth}{!}{$\displaystyle
\underbrace{
\frac{\lambda}{|t_{\mathrm{s}}|}
\sum_{t=1}^{|t_{\mathrm{s}}|}
\mathcal{C}(r_{i,t}^{s}, \hat{A}_{i,t}^{s})
+
\frac{\lambda}{|o_i|-|t_{\mathrm{s}}|}
\sum_{t=|t_{\mathrm{s}}|+1}^{|o_i|}
\mathcal{C}(r_{i,t}^{d}, \hat{A}_{i,t}^{d})
}_{\substack{\text{local clipped surrogate objective}}}
$}
\\
&\resizebox{\columnwidth}{!}{$\displaystyle
+
\underbrace{
\frac{1}{|o_i|}
\sum_{t=1}^{|o_i|}
\Bigg(
\mathcal{C}(r_{i,t}^{g}, \hat{A}_{i,t}^{g})
}_{\substack{\text{global clipped surrogate objective}}}
-
\beta\operatorname{KL}\!\left(
\pi_{\theta} \,\|\, \pi_{\mathrm{ref}}
\right)
\Bigg)
\Bigg]
\Bigg\}
$}
\\
&=
\mathbb{E}_{\substack{
(q,a) \sim \mathcal{D} \\
\{o_i\}_{i=1}^{G} \sim \pi_{\theta_{\mathrm{old}}}
}}
\Bigg\{
\frac{1}{G}\sum_{i=1}^{G}
\Bigg[
\\
&\sum_{t=1}^{|t_{\mathrm{s}}|}
\Bigg(
\frac{\lambda}{|t_{\mathrm{s}}|}
\mathcal{C}\!\left(r_{i,t}^{s}, \hat{A}_{i,t}^{s}\right)
+
\frac{1}{|o_i|}
\mathcal{C}\!\left(r_{i,t}^{g}, \hat{A}_{i,t}^{g}\right)
\\
&-
\frac{\beta}{|o_i|}
\operatorname{KL}\!\left(
\pi_{\theta}(\cdot \mid q,o_{i,<t})
\,\|\,\pi_{\mathrm{ref}}(\cdot \mid q,o_{i,<t})
\right)
\Bigg)
\\
&+
\sum_{t=|t_{\mathrm{s}}|+1}^{|o_i|}
\Bigg(
\frac{\lambda}{|o_i|-|t_{\mathrm{s}}|}
\mathcal{C}\!\left(r_{i,t}^{d}, \hat{A}_{i,t}^{d}\right)
\\
&+
\frac{1}{|o_i|}
\mathcal{C}\!\left(r_{i,t}^{g}, \hat{A}_{i,t}^{g}\right)
-
\frac{\beta}{|o_i|}
\operatorname{KL}\!\left(
\pi_{\theta}(\cdot \mid q,o_{i,<t})
\,\|\,\right.
\\
&\left.
\pi_{\mathrm{ref}}(\cdot \mid q,o_{i,<t})
\right)
\Bigg)
\Bigg]
\Bigg\}.
\end{aligned}
\end{equation}

This time-wise form makes the effective credit assignment explicit: before
the boundary $|t_{\text{s}}|$, each token is optimized with a streaming-local
coefficient $\lambda/|t_{\text{s}}|$ and a global coefficient $1/|o_i|$; after
the boundary, the streaming-local term is replaced by the deep-local
coefficient $\lambda/(|o_i|\!-\!|t_{\text{s}}|)$.

We next derive the corresponding policy gradient. By linearity of
differentiation and expectation,
\begin{equation}
\label{eq:hrpo_gradient_expanded}
\begin{aligned}
&\nabla_{\theta}J_{\mathrm{HRPO}}(\theta)
=
\mathbb{E}_{\substack{
(q,a) \sim \mathcal{D} \\
\{o_i\}_{i=1}^{G} \sim \pi_{\theta_{\mathrm{old}}}
}}
\Bigg\{
\frac{1}{G}\sum_{i=1}^{G}
\Bigg[
\\
&\frac{\lambda}{|t_{\mathrm{s}}|}
\sum_{t=1}^{|t_{\mathrm{s}}|}
\nabla_{\theta}\mathcal{C}\!\left(
r_{i,t}^{s}, \hat{A}_{i,t}^{s}
\right)
\\
&+
\frac{\lambda}{|o_i|-|t_{\mathrm{s}}|}
\sum_{t=|t_{\mathrm{s}}|+1}^{|o_i|}
\nabla_{\theta}\mathcal{C}\!\left(
r_{i,t}^{d}, \hat{A}_{i,t}^{d}
\right)
\\
&+
\frac{1}{|o_i|}
\sum_{t=1}^{|o_i|}
\nabla_{\theta}\mathcal{C}\!\left(
r_{i,t}^{g}, \hat{A}_{i,t}^{g}
\right)
\\
&-
\frac{\beta}{|o_i|}
\sum_{t=1}^{|o_i|}
\nabla_{\theta}
\operatorname{KL}\!\left(
\right.
\\
&\left.
\pi_{\theta}(\cdot \mid q,o_{i,<t})
\,\|\, 
\pi_{\mathrm{ref}}(\cdot \mid q,o_{i,<t})
\right)
\Bigg]
\Bigg\}.
\end{aligned}
\end{equation}
For each $\ell\in\{s,d,g\}$, the sampled
advantage $\hat{A}_{i,t}^{\ell}$ and the old-policy denominator in
$r_{i,t}^{\ell}$ are fixed with respect to $\theta$, so
\begin{equation}
    \nabla_{\theta}r_{i,t}^{\ell}
    =
    r_{i,t}^{\ell}
    \nabla_{\theta}\log\pi_{\theta}(o_{i,t}|q,o_{i,<t}).
    \label{eq:ratio_gradient}
\end{equation}
Therefore the clipped surrogate contributes a policy-gradient term only when
the unclipped PPO branch is selected:
\begin{equation}
\label{eq:clipped_surrogate_gradient}
\begin{aligned}
&\nabla_{\theta}\mathcal{C}\!\left(
r_{i,t}^{\ell},\,\hat{A}_{i,t}^{\ell}
\right)
=
\\
&\begin{cases}
\begin{aligned}
&r_{i,t}^{\ell}\hat{A}_{i,t}^{\ell}
\nabla_{\theta}\log \pi_{\theta}
\\
&\left(
o_{i,t}\mid q,o_{i,<t}
\right),
\end{aligned}
&
\begin{aligned}
&\hat{A}_{i,t}^{\ell}\ge 0
\ \text{and}\ 
r_{i,t}^{\ell}\le 1+\varepsilon,\\[-2pt]
&\text{or }\hat{A}_{i,t}^{\ell}<0
\ \text{and}\ 
\\
&r_{i,t}^{\ell}\ge 1-\varepsilon
\end{aligned}
\\
0, & \text{otherwise}.
\end{cases}
\end{aligned}
\end{equation}
Substituting Eq.~\ref{eq:clipped_surrogate_gradient} into
Eq.~\ref{eq:hrpo_gradient_expanded} gives the time-wise policy gradient:
{
\begin{equation}
\label{eq:hrpo_policy_gradient}
\begin{aligned}
&\nabla_{\theta}J_{\mathrm{HRPO}}(\theta)
=
\mathbb{E}_{\substack{
(q,a) \sim \mathcal{D} \\
\{o_i\}_{i=1}^{G} \sim \pi_{\theta_{\mathrm{old}}}
}}
\Bigg\{
\frac{1}{G}\sum_{i=1}^{G}
\Bigg[
\\[-2pt]
&\sum_{t=1}^{|t_{\mathrm{s}}|}
\Bigg(
\frac{\lambda}{|t_{\mathrm{s}}|}
m_{i,t}^{s} r_{i,t}^{s} \hat{A}_{i,t}^{s}
+
\frac{1}{|o_i|}
m_{i,t}^{g} r_{i,t}^{g} \hat{A}_{i,t}^{g}
\Bigg)
\\
&\nabla_{\theta}\log
\pi_{\theta}\!\left(
o_{i,t} \mid q,o_{i,<t}
\right)
\\
&-
\sum_{t=1}^{|t_{\mathrm{s}}|}
\frac{\beta}{|o_i|}
\nabla_{\theta}
\operatorname{KL}\!\left(
\pi_{\theta}(\cdot \mid q,o_{i,<t})
\,\|\,
\right.
\\
&\left.
\pi_{\mathrm{ref}}(\cdot \mid q,o_{i,<t})
\right)
\\
&+
\sum_{t=|t_{\mathrm{s}}|+1}^{|o_i|}
\Bigg(
\frac{\lambda}{|o_i|-|t_{\mathrm{s}}|}
m_{i,t}^{d} r_{i,t}^{d} \hat{A}_{i,t}^{d}
\\
&\qquad
+
\frac{1}{|o_i|}
m_{i,t}^{g} r_{i,t}^{g} \hat{A}_{i,t}^{g}
\Bigg)
\\
&\nabla_{\theta}\log
\pi_{\theta}\!\left(
o_{i,t} \mid q,o_{i,<t}
\right)
\\
&-
\sum_{t=|t_{\mathrm{s}}|+1}^{|o_i|}
\frac{\beta}{|o_i|}
\nabla_{\theta}
\operatorname{KL}\!\left(
\right.
\\
&\left.
\pi_{\theta}(\cdot \mid q,o_{i,<t})
\,\|\, 
\pi_{\mathrm{ref}}(\cdot \mid q,o_{i,<t})
\right)
\Bigg]
\Bigg\}.
\end{aligned}
\end{equation}
}
where $m_{i,t}^{\ell}$ is the binary condition in the first branch of
Eq.~\ref{eq:clipped_surrogate_gradient}; it is not a new reward or advantage
variable, but only records whether the corresponding clipped surrogate is
active. Equivalently, if no clipping is active at token $t$, the streaming-token
and deep-token gradient factors reduce to
{
\begin{equation}
\label{eq:hrpo_streaming_gradient_term}
\begin{aligned}
&
\left(
\frac{\lambda}{|t_{\text{s}}|}
r_{i,t}^s \hat{A}_{i,t}^s
+
\frac{1}{|o_i|}
r_{i,t}^g \hat{A}_{i,t}^g
\right)
\\
&\cdot
\nabla_{\theta}\log
\pi_{\theta}\!\left(
o_{i,t} \mid q,o_{i,<t}
\right),
\text{for } 1 \le t \le |t_{\text{s}}| .
\end{aligned}
\end{equation}
}

and

{
\begin{equation}
\label{eq:hrpo_deep_gradient_term}
\begin{aligned}
&
\left(
\frac{\lambda}{|o_i|-|t_{\text{s}}|}
r_{i,t}^d \hat{A}_{i,t}^d
+
\frac{1}{|o_i|}
r_{i,t}^g \hat{A}_{i,t}^g
\right)
\\
&\cdot
\nabla_{\theta}\log
\pi_{\theta}\!\left(
o_{i,t} \mid q,o_{i,<t}
\right),
\text{for } |t_{\text{s}}| < t \le |o_i| .
\end{aligned}
\end{equation}
}
with the KL-gradient term subtracted at every token as shown in
Eq.~\ref{eq:hrpo_policy_gradient}. At $\theta=\theta_{\mathrm{old}}$, where
the ratios equal one, HRPO uses the effective per-token advantages
\begin{equation}
    \frac{\lambda}{|t_{\text{s}}|}\hat{A}_{i,t}^s
    +\frac{1}{|o_i|}\hat{A}_{i,t}^g
    , \quad 
    \frac{\lambda}{|o_i|\!-\!|t_{\text{s}}|}\hat{A}_{i,t}^d
    +\frac{1}{|o_i|}\hat{A}_{i,t}^g
\end{equation}
for streaming and deep tokens respectively.

\section{Reward Analysis for Streaming Reasoning}

\paragraph{Adaptive reward hacking}
\label{app:adaptive-reward-hacking}
An intuitive design for local length reward is to add accuracy and format
gates, so that only correct and parseable trajectories receive the local
length signal. However, our local length term is actually a \emph{penalty}.
Consider $\ell_i^s\le 0$ and the gated design
$R_i^{L_s}=R_i^{\mathrm{acc}}R_{i,\mathrm{fmt}}^s\ell_i^s$. Correct and
format-valid samples are penalized when they are too long, while incorrect or
malformed samples are multiplied by the gate and therefore receive $0$ penalty.
The gate thus creates a loophole: failing the gate can be better than being
correct but verbose.
\begin{center}
\begin{tabular}{lccc}
\hline
Rollout & Accuracy & Format & $R_i^{L_s}$ \\
\hline
A: correct but long & 1 & 1 & $-3$ \\
B: incorrect & 0 & 1 & $0$ \\
C: incorrect & 0 & 1 & $0$ \\
D: incorrect & 0 & 1 & $0$ \\
\hline
\end{tabular}
\end{center}

After group normalization, the zero-valued incorrect rollouts can obtain
positive local length advantages, while the correct long rollout obtains a
negative one. The local branch can therefore reward failed trajectories for
avoiding the length cost, creating a direct conflict with the global signal.

This failure may appear to be solvable by carefully tuning the coefficient of
the local length reward. In the gated regime, however, weighting the reward
before computing the advantage cancels the coefficient during normalization, as
shown in Eq.~\ref{eq:reward_level_beta_cancels}. Therefore, local length scores
should not be gated by final accuracy, and format validity should be handled by
masking or filtering at the advantage-computation level rather than by
multiplying the negative penalty itself.

\paragraph{Hyperparameter Sensitivity Analysis}
\label{app:beta-sensitivity}

We analyze why HRPO applies $\beta$ after component-wise normalization in
Eq.~\ref{eq:advantage_composition}. Consider a reward-level alternative at any
hierarchical level $\ell\in\{s,d,g\}$:
\begin{equation}
    R_i^\ell(\beta)=b^\ell+\beta g^\ell X_i^\ell,\qquad \beta>0,
\end{equation}
where $b^\ell$ is the shared additive gate reward, $g^\ell>0$ is the shared
multiplicative gate scale, and $X_i^\ell$ is the remaining length or efficiency
shaping term. In the active gated regime, $b^\ell$ and $g^\ell$ are fixed within
the sampled group. The normalized advantage is therefore:
\begin{equation}
\label{eq:reward_level_beta_cancels}
\begin{aligned}
\hat{A}_i^\ell(\beta)
&=
\frac{
R_i^\ell(\beta)
-
\mu\big(\{R_j^\ell(\beta)\}_{j=1}^G\big)
}{
\sigma\big(\{R_j^\ell(\beta)\}_{j=1}^G\big)
}
\\
&=
\frac{
\beta g^\ell\!\left(
X_i^\ell-\mu(\{X_j^\ell\}_{j=1}^G)
\right)
}{
\beta g^\ell
\sigma(\{X_j^\ell\}_{j=1}^G)
}
\\
&=
\frac{
X_i^\ell-\mu(\{X_j^\ell\}_{j=1}^G)
}{
\sigma(\{X_j^\ell\}_{j=1}^G)
}.
\end{aligned}
\end{equation}

Thus reward-level weighting makes $\beta$ ineffective under fixed gates. In
contrast, Eq.~\ref{eq:advantage_composition} applies $\beta$ after
normalization, so it remains an explicit multiplier on the composed advantage.
If a component has zero variance within the group, that component still
produces no relative advantage.
The corresponding experiments can be found at Table~\ref{tab:hyperparameters}.

\paragraph{Hyperparameter Weighting Experiments}
\begin{table*}[t]
\centering
\footnotesize
\setlength{\tabcolsep}{3pt}
\renewcommand{\arraystretch}{0.88}
\resizebox{\textwidth}{!}{%
\begin{tabular}{@{}ccccc l rrrrrr@{}}
\toprule
\multicolumn{2}{c}{Reward coefficients} & \multirow{2}{*}{$\alpha$} & \multirow{2}{*}{$\beta$} & \multirow{2}{*}{$\lambda$} & \multirow{2}{*}{Method}
& \multicolumn{2}{c}{GSM-symbolic P2}
& \multicolumn{2}{c}{GSM-symbolic P1}
& \multicolumn{2}{c}{MetaMathQA} \\
\cmidrule(lr){1-2}\cmidrule(lr){7-8}\cmidrule(lr){9-10}\cmidrule(l){11-12}
Acc. & Format & & & & & Acc & Len. & Acc & Len. & Acc & Len. \\
\midrule
\multicolumn{12}{@{}l}{\textit{Baselines}} \\
/ & / & 0.5 & / & / & SFT & 0.642 & 356.910 & 0.839 & 286.132 & 0.688 & 270.661 \\
1 & / & 0.5 & / & / & GRPO & 0.726 & 392.908 & 0.877 & 301.709 & 0.805 & 297.310 \\
\midrule
\multicolumn{12}{@{}l}{\textit{Format reward and local-global balance without adaptive length reward}} \\
1 & 2 & 0.5 & / & 0.1 & HRPO & 0.782 & 498.728 & 0.895 & 377.679 & 0.842 & 385.627 \\
1 & 2 & 0.5 & / & 0.5 & HRPO & 0.774 & 452.712 & 0.886 & 341.994 & 0.816 & 371.474 \\
1 & 2 & 0.5 & / & 1 & HRPO & 0.758 & 381.932 & 0.872 & 314.082 & 0.802 & 304.650 \\
1 & 2 & 0.5 & / & 0.5 & HRPO-sentence & 0.798 & 436.818 & 0.926 & 346.441 & 0.858 & 337.687 \\
1 & 2 & 0.5 & / & 0.5 & HRPO-token & 0.776 & 435.020 & 0.888 & 330.450 & 0.815 & 343.115 \\
1 & 2 & 0.5 & / & 1 & HRPO & 0.756 & 377.750 & 0.869 & 323.723 & 0.804 & 321.838 \\
1 & 2 & 0.5 & / & 1 & HRPO-sentence & 0.736 & 433.320 & 0.879 & 341.503 & 0.825 & 342.731 \\
1 & 2 & 0.5 & / & 0.05 & HRPO & 0.800 & 433.080 & 0.884 & 331.890 & 0.817 & 343.084 \\
\midrule
\multicolumn{12}{@{}l}{\textit{Adaptive length-reward weight $\beta$ at fixed $\lambda=0.5$}} \\
1 & 2 & 0.5 & 0.05 & 0.5 & HRPO & 0.642 & 442.900 & 0.818 & 315.474 & 0.655 & 461.854 \\
1 & 2 & 0.5 & 0.1 & 0.5 & HRPO & 0.532 & 322.770 & 0.669 & 403.125 & 0.583 & 531.450 \\
1 & 2 & 0.5 & 0.5 & 0.5 & HRPO & 0.134 & 610.698 & 0.195 & 630.160 & 0.293 & 621.358 \\
1 & 2 & 0.5 & 1 & 0.5 & HRPO & 0.088 & 136.512 & 0.154 & 164.506 & 0.320 & 173.537 \\
\midrule
\multicolumn{12}{@{}l}{\textit{Granularity comparison with stronger local weighting ($\lambda=1$)}} \\
1 & 2 & 0.5 & 0.05 & / & GRPO & 0.054 & 73.670 & 0.083 & 73.890 & 0.194 & 89.980 \\
1 & 2 & 0.5 & 0.05 & 1 & HRPO & 0.620 & 316.740 & 0.770 & 273.110 & 0.664 & 334.560 \\
1 & 2 & 0.5 & 0.05 & 1 & HRPO-sentence & 0.596 & 402.470 & 0.621 & 629.650 & 0.565 & 701.251 \\
1 & 2 & 0.5 & 0.05 & 1 & HRPO-token & 0.450 & 240.300 & 0.635 & 173.960 & 0.662 & 188.980 \\
\midrule
\multicolumn{12}{@{}l}{\textit{Local-global objective balance $\lambda$ with adaptive length reward}} \\
1 & 2 & 0.5 & 0.05 & 0.01 & HRPO & 0.776 & 384.980 & 0.874 & 299.130 & 0.813 & 307.040 \\
1 & 2 & 0.5 & 0.05 & 0.05 & HRPO & 0.750 & 373.060 & 0.870 & 290.530 & 0.801 & 293.200 \\
1 & 2 & 0.5 & 0.05 & 0.1 & HRPO & 0.760 & 357.980 & 0.869 & 279.850 & 0.810 & 272.600 \\
1 & 2 & 0.5 & 0.05 & 1 & HRPO & 0.620 & 316.740 & 0.770 & 273.110 & 0.664 & 334.560 \\
\midrule
\multicolumn{12}{@{}l}{\textit{Default setting and latency-discount sensitivity}} \\
1 & 2 & 0.5 & 0.05 & / & GRPO & 0.758 & 384.488 & 0.85 & 306.792 & 0.823 & 317.997 \\
1 & 2 & 0.5 & 0.05 & 0.05 & HRPO & 0.788 & 370.256 & 0.871 & 292.004 & 0.826 & 302.602 \\
1 & 2 & 0.75 & 0.05 & 0.05 & HRPO & 0.800 & 353.230 & 0.887 & 291.355 & 0.775 & 276.256 \\
1 & 2 & 0.25 & 0.05 & 0.05 & HRPO & 0.784 & 418.572 & 0.894 & 311.095 & 0.821 & 344.982 \\
1 & 2 & 1 & 0.05 & 0.05 & HRPO & 0.794 & 351.022 & 0.876 & 291.810 & 0.810 & 270.148 \\
\midrule
\multicolumn{12}{@{}l}{\textit{Fine-grained variants under the default reward setting}} \\
1 & 2 & 0.5 & 0.05 & 0.05 & HRPO-sentence & 0.762 & 381.170 & 0.873 & 296.516 & 0.809 & 300.954 \\
1 & 2 & 0.5 & 0.05 & 0.05 & HRPO-token & 0.756 & 354.008 & 0.860 & 278.407 & 0.805 & 276.448 \\
\midrule
\multicolumn{12}{@{}l}{\textit{Format reward coefficient sensitivity at fixed Acc. coefficient $=1$}} \\
1 & 1 & 0.5 & 0.05 & 0.05 & HRPO& 0.772 & 361.840 & 0.858 & 284.910 & 0.807 & 294.730 \\
1 & 2 & 0.5 & 0.05 & 0.05 & HRPO & 0.788 & 370.256 & 0.871 & 292.004 & 0.826 & 302.602 \\
1 & 5 & 0.5 & 0.05 & 0.05 & HRPO& 0.764 & 405.620 & 0.852 & 321.780 & 0.798 & 335.440 \\
\bottomrule
\end{tabular}
}
\caption{Hyperparameter settings and evaluation results for Qwen3-1.7B. The first two columns report the coefficients of the accuracy and format rewards, while $\alpha$ is the latency discount, $\beta$ is the adaptive length-reward weight, and $\lambda$ controls the local-global balance in HRPO. Rows are grouped by ablation theme, and evaluation reports accuracy and total generation length on GSM-symbolic P1/P2 and MetaMathQA.}
\label{tab:hyperparameters}
\end{table*}

We investigate the role of different hyperparameters in both reward composition and advantage weighting, with the results summarized in Table ~\ref{tab:hyperparameters}. Specifically, we study the effects of the format, length, and accuracy reward weights, as well as the advantage weighting coefficient $\lambda$, on GRPO, HRPO, and the finer-grained HRPO-Sentence and HRPO-Token variants using Qwen3-1.7B. Based on these comparisons, we select the configuration with Acc: 1, format: 2, length: 0.05, and $\lambda=0.05$ as the default setting in our paper, since it provides a relatively robust trade-off among accuracy, formatting reliability, and response length.

\section{Evaluation Details of AdaSR}
\paragraph{Multiseed Results with Standard Deviation}

\begin{table*}[t]
\centering
\scriptsize
\setlength{\tabcolsep}{4.0pt}
\renewcommand{\arraystretch}{0.92}
\resizebox{\textwidth}{!}{%
\begin{tabular}{lllcccc}
\toprule
Model & Benchmark & Metric
& Read-then-think
& StreamingThinker (SFT)~\citep{tong2025streamingthinker}
& AdaSR-GRPO (Ours)
& AdaSR-HRPO (Ours) \\
\midrule

\multirow{6}{*}{\rotatebox[origin=c]{90}{\textit{Qwen3-1.7B}}}
& \multirow{2}{*}{GSM-symbolic P1}
& Acc$\uparrow$ & 0.634 & 0.834
& 0.863 {\scriptsize $\pm$ 0.0087}
& \textbf{0.872 {\scriptsize $\pm$ 0.0041}} \\
& & Total Len.$\downarrow$ & 1598.699 & \textbf{283.749}
& 305.108 {\scriptsize $\pm$ 2.973}
& \textbf{289.930 {\scriptsize $\pm$ 1.392}} \\
\cmidrule(lr){2-7}

& \multirow{2}{*}{GSM-symbolic P2}
& Acc$\uparrow$ & 0.424 & 0.642
& 0.765 {\scriptsize $\pm$ 0.0074}
& \textbf{0.773 {\scriptsize $\pm$ 0.0135}} \\
& & Total Len.$\downarrow$ & 1866.474 & \textbf{361.906}
& 387.673 {\scriptsize $\pm$ 3.602}
& \textbf{373.438 {\scriptsize $\pm$ 2.568}} \\
\cmidrule(lr){2-7}

& \multirow{2}{*}{MetaMathQA}
& Acc$\uparrow$ & 0.848 & 0.688
& \textbf{0.818 {\scriptsize $\pm$ 0.0090}}
& 0.815 {\scriptsize $\pm$ 0.0140} \\
& & Total Len.$\downarrow$ & 1281.774 & 270.661
& 318.163 {\scriptsize $\pm$ 1.751}
& \textbf{296.808 {\scriptsize $\pm$ 8.091}} \\

\midrule

\multirow{6}{*}{\rotatebox[origin=c]{90}{\textit{Qwen3-4B}}}
& \multirow{2}{*}{GSM-symbolic P1}
& Acc$\uparrow$ & 0.636 & 0.934
& \textbf{0.943 {\scriptsize $\pm$ 0.0043}}
& 0.935 {\scriptsize $\pm$ 0.0104} \\
& & Total Len.$\downarrow$ & 1717.155 & 286.992
& 268.948 {\scriptsize $\pm$ 0.727}
& \textbf{257.665 {\scriptsize $\pm$ 6.439}} \\
\cmidrule(lr){2-7}

& \multirow{2}{*}{GSM-symbolic P2}
& Acc$\uparrow$ & 0.462 & 0.816
& 0.871 {\scriptsize $\pm$ 0.0055}
& \textbf{0.876 {\scriptsize $\pm$ 0.0112}} \\
& & Total Len.$\downarrow$ & 1937.396 & 357.584
& 336.332 {\scriptsize $\pm$ 1.203}
& \textbf{317.438 {\scriptsize $\pm$ 6.752}} \\
\cmidrule(lr){2-7}

& \multirow{2}{*}{MetaMathQA}
& Acc$\uparrow$ & 0.828 & 0.860
& 0.903 {\scriptsize $\pm$ 0.0050}
& \textbf{0.909 {\scriptsize $\pm$ 0.0128}} \\
& & Total Len.$\downarrow$ & 1379.088 & 274.198
& 251.699 {\scriptsize $\pm$ 1.518}
& \textbf{237.388 {\scriptsize $\pm$ 0.618}} \\

\bottomrule
\end{tabular}
}
\caption{Main multiseed results on streaming reasoning benchmarks with Qwen3-1.7B and Qwen3-4B.
For AdaSR-GRPO and AdaSR-HRPO.}
\label{tab:main_std}
\end{table*}

\begin{table*}[t]
\centering
\setlength{\tabcolsep}{5.0pt}
\renewcommand{\arraystretch}{0.92}
\resizebox{\textwidth}{!}{%
\begin{tabular}{lcccccc}
\toprule
Qwen3-1.7B
& \multicolumn{2}{c}{GSM-symbolic P2}
& \multicolumn{2}{c}{GSM-symbolic P1}
& \multicolumn{2}{c}{MetaMathQA} \\
\cmidrule(lr){2-3}
\cmidrule(lr){4-5}
\cmidrule(lr){6-7}
Method
& Acc$\uparrow$ & Total Len.$\downarrow$
& Acc$\uparrow$ & Total Len.$\downarrow$
& Acc$\uparrow$ & Total Len.$\downarrow$ \\
\midrule
GRPO
& 0.765 {\scriptsize $\pm$ 0.0074} & 387.673 {\scriptsize $\pm$ 3.602}
& 0.867 {\scriptsize $\pm$ 0.0024} & 305.108 {\scriptsize $\pm$ 2.973}
& \textbf{0.818 {\scriptsize $\pm$ 0.0090}} & 318.163 {\scriptsize $\pm$ 1.751} \\
\noalign{\vskip 1pt}
\cdashline{1-7}
\noalign{\vskip 1pt}
HRPO
& \textbf{0.773 {\scriptsize $\pm$ 0.0135}} & 373.438 {\scriptsize $\pm$ 2.568}
& \textbf{0.872 {\scriptsize $\pm$ 0.0041}} & 289.930 {\scriptsize $\pm$ 1.392}
& 0.815 {\scriptsize $\pm$ 0.0140} & 296.808 {\scriptsize $\pm$ 8.091} \\
HRPO-sentence
& 0.757 {\scriptsize $\pm$ 0.0060} & 383.475 {\scriptsize $\pm$ 1.352}
& 0.870 {\scriptsize $\pm$ 0.0039} & 296.344 {\scriptsize $\pm$ 0.764}
& 0.807 {\scriptsize $\pm$ 0.0071} & 306.744 {\scriptsize $\pm$ 1.029} \\
HRPO-token
& 0.764 {\scriptsize $\pm$ 0.0050} & \textbf{356.935 {\scriptsize $\pm$ 1.643}}
& 0.862 {\scriptsize $\pm$ 0.0047} & \textbf{279.256 {\scriptsize $\pm$ 1.681}}
& 0.801 {\scriptsize $\pm$ 0.0012} & \textbf{277.332 {\scriptsize $\pm$ 0.860}} \\
\bottomrule
\end{tabular}
}
\caption{Multiseed results of hierarchical advantage assignment on Qwen3-1.7B.
We report mean $\pm$ sample standard deviation over available runs.}
\label{tab:hrpo_std}
\end{table*}

To assess the robustness of our results, we conduct experiments under multiple random seeds and report the mean performance together with the corresponding standard deviation. The detailed results are presented in Table~\ref{tab:main_std} and Table~\ref{tab:hrpo_std}. 
\paragraph{Out-of-Domain Experiment Details}
In Table~\ref{tab:ood_gsm_infinite_logicnli}, our SFT-base model is trained on the math-based datasets P1, P2, and MetaMathQA. GSM-Infinite is an out-of-domain mathematical benchmark, while LogicNLI is a logic-based benchmark used to evaluate the model’s out-of-task generalization ability.
\section{Training Details of AdaSR}
\label{app:training-details}

\begin{figure*}[!h]
    \centering
    \includegraphics[width=0.95\textwidth]{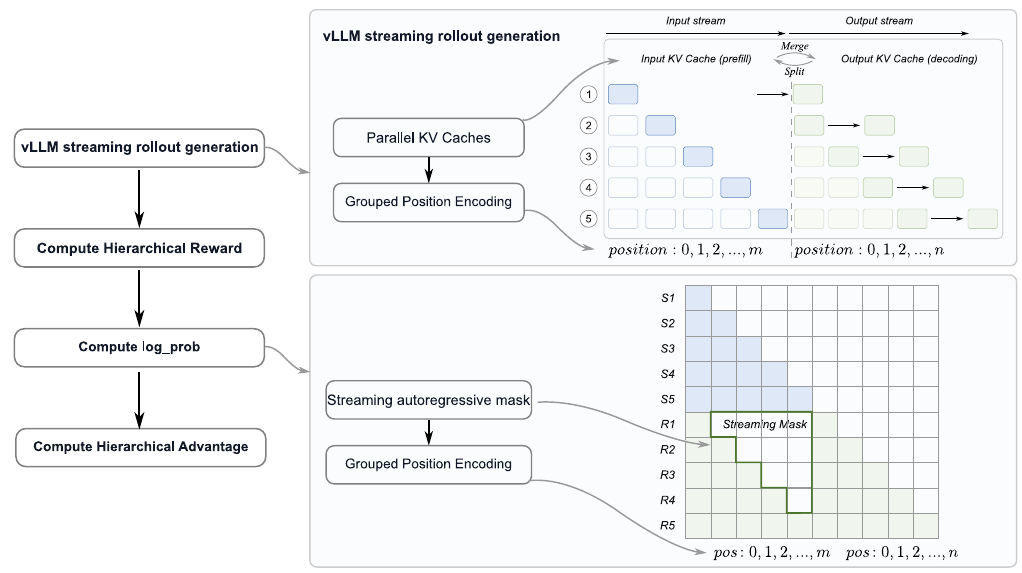}
    \caption{Illustration of the AdaSR training pipeline. The left panel shows the overall training pipeline, where streaming rollouts are generated, hierarchical rewards are computed, log probabilities are evaluated, and hierarchical advantages are assigned. The upper-right panel illustrates the rollout-side implementation, where parallel input/output KV caches and grouped position encoding enable streaming generation with separated prefill and decoding states. The lower-right panel shows the log-probability computation, where a streaming autoregressive mask and grouped position encoding preserve the same partial-observation structure during policy evaluation. Together, these components support consistent streaming rollout generation and streaming-aware hierarchical advantage computation.}
    \label{fig:rollout}
\end{figure*}

While the main text provides the algorithmic workflow of AdaSR, this appendix further details the concrete streaming rollout and stream-aware logits computation, as illustrated in Figure~\ref{fig:rollout}.

\begin{table}[t]
\centering
\footnotesize
\setlength{\tabcolsep}{3pt}
\renewcommand{\arraystretch}{1.05}
\begin{tabular}{@{}p{0.62\columnwidth}p{0.28\columnwidth}@{}}
\toprule
\textbf{Parameter} & \textbf{Value} \\
\midrule
Weight of Local clipped surrogate term ($\lambda$) & $0.05$\\
Weight of Local Length Advantage ($\beta$) & $0.05$ \\
Latency discount coefficient $\alpha$ & 0.5\\
scale coefficient $\tau$ & Avg of Token Length\\
Train batch size & 16\\
PPO mini batch size & $16$ \\
PPO micro batch size per GPU & $2$ \\
Log-prob micro batch size per GPU & $8$ \\
KL coefficient & False \\
Max prompt length & $8192$ tokens \\
Max response length & $2048$ tokens \\
Max model length & $4096$ tokens \\
Sampling temperature & $0.6$ \\
Top-$p$ sampling & $0.95$ \\
Top-$k$ sampling & $20$ \\
Number of samples per prompt & $12$ \\
Intermediate max tokens & $512$ tokens \\
Compression ratio & $0.35$ \\
Minimum reasoning tokens & $20$ tokens \\
Maximum reasoning tokens & $2048$ tokens \\
Learning rate & 2e-6\\
Weight decay & 0.01 \\
Warmup ratio & 0.05\\
\bottomrule
\end{tabular}
\caption{Hyperparameters in streaming RL training.}
\label{tab:RL_hyperparameters}
\end{table}

\paragraph{Procedure of AdaSR algorithm}
Algorithm~\ref{alg:adasr} summarizes the overall training procedure of AdaSR. Starting from a streaming SFT policy, AdaSR performs group sampling for each input, generates sentence-level streaming rationales followed by a deep reasoning stage, and then computes reward components for each sampled trajectory. These rewards are further decomposed into streaming, deep, and global advantages, which are assigned to the corresponding token groups through the hierarchical advantage mechanism. Finally, the policy is optimized with the HRPO objective, allowing the model to improve answer accuracy while explicitly controlling the contribution of different reasoning stages.

\paragraph{Hyperparameters}

We initialize AdaSR from the corresponding StreamingThinker SFT checkpoint and optimize it with the streaming RL stack described below. Unless otherwise specified, the actor and reference policy are trained with FSDP, while rollout generation is served by vLLM~\citep{kwon2023efficient}. In our main implementation, vLLM rollout uses tensor parallelism over four GPUs. All experiments are conducted on A100 GPUs. The detailed parameters are shown in the Table~\ref{tab:RL_hyperparameters}.

\paragraph{veRL Adaptation}
AdaSR is implemented by adapting the standard verl training loop
\citep{sheng2024hybridflow}. The original verl pipeline assumes that rollout
prompts are fully available before generation and that actor log-probability computation uses a standard full-context causal mask. This assumption is incompatible with streaming reasoning, where each streaming segment must be generated and scored under a partial-observation constraint. We therefore replace the default trainer, rollout engine, and actor forward path at runtime:
\texttt{RayPPOTrainer} is replaced by a streaming-aware trainer,
\texttt{vLLMRollout} by \texttt{StreamingThinkerVLLMRollout}, and the actor
model by a Qwen3 forward implementation that accepts streaming metadata. This keeps verl's distributed PPO/GRPO infrastructure, FSDP actor updates, reference policy evaluation, logging, and checkpointing intact, while changing only the parts that define rollout semantics and token visibility.

The training order is important. AdaSR first computes rewards on the raw rollout batch, because reward functions need the generated text and the round-level streaming/deep segment boundaries. Only after reward computation do we repack the rollout into the StreamingThinker training layout. This repacking step
reconstructs the source and target segment lists, preserves the generated
round boundaries, and writes metadata such as \texttt{\_lengths},
\texttt{source\_seg\_len}, \texttt{target\_seg\_len}, and
\texttt{target\_seg\_roles}. The repacked batch is then used for old
log-probability, reference log-probability, advantage assignment, and actor
update, ensuring that all policy-gradient quantities are computed under the
same streaming visibility.

\paragraph{Rollout Generation}

\begin{figure*}[!h]
    \centering
    \includegraphics[width=0.95\textwidth]{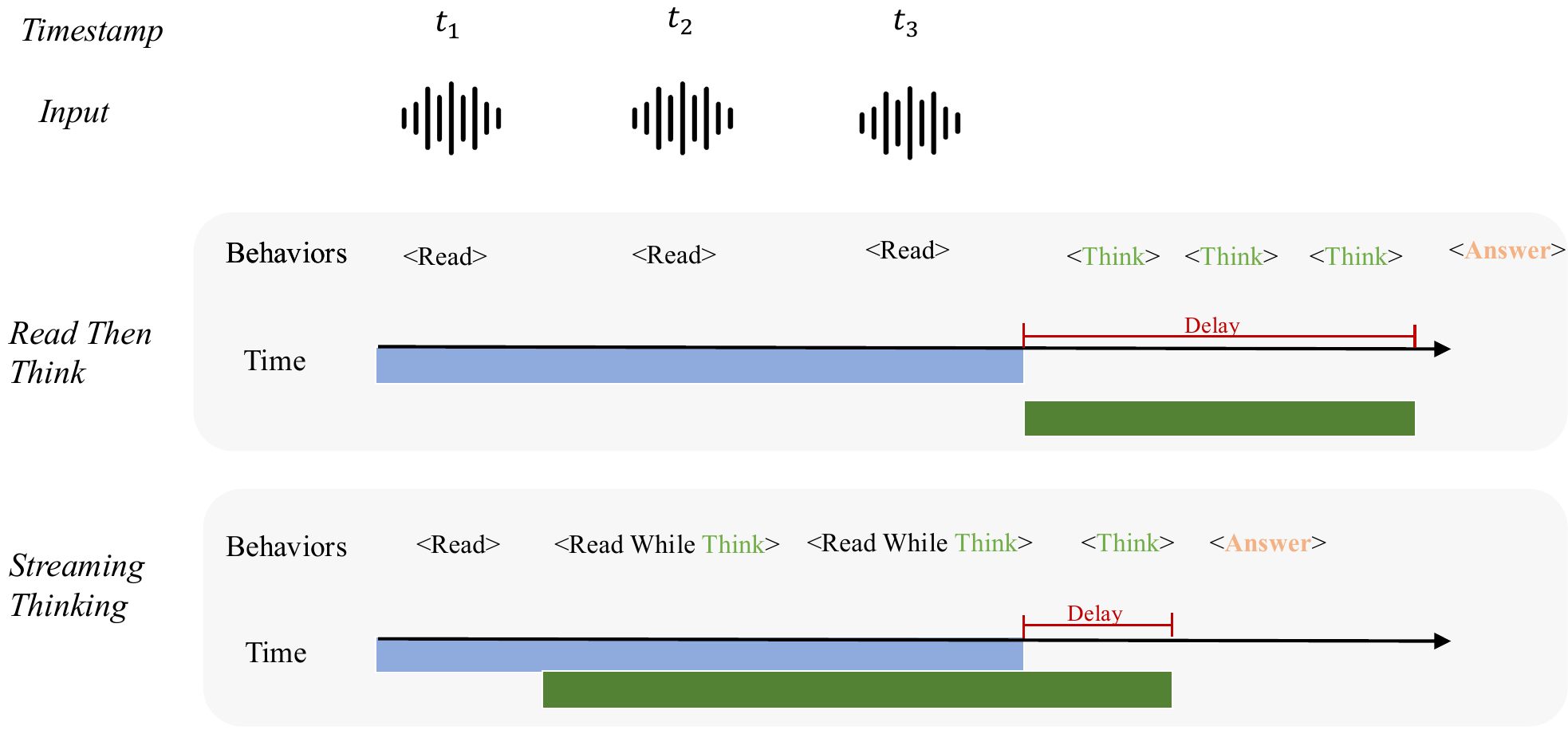}
    \caption{Comparison of reasoning paradigms. Streaming thinking performs reasoning concurrently with incremental input reading, enabling earlier responses and reducing delay. In contrast, the read-then-think paradigm postpones reasoning until all inputs are received, resulting in a larger delay.}
    \label{fig:latency}
\end{figure*}
\begin{figure*}[!h]
    \centering
    \includegraphics[width=0.95\textwidth]{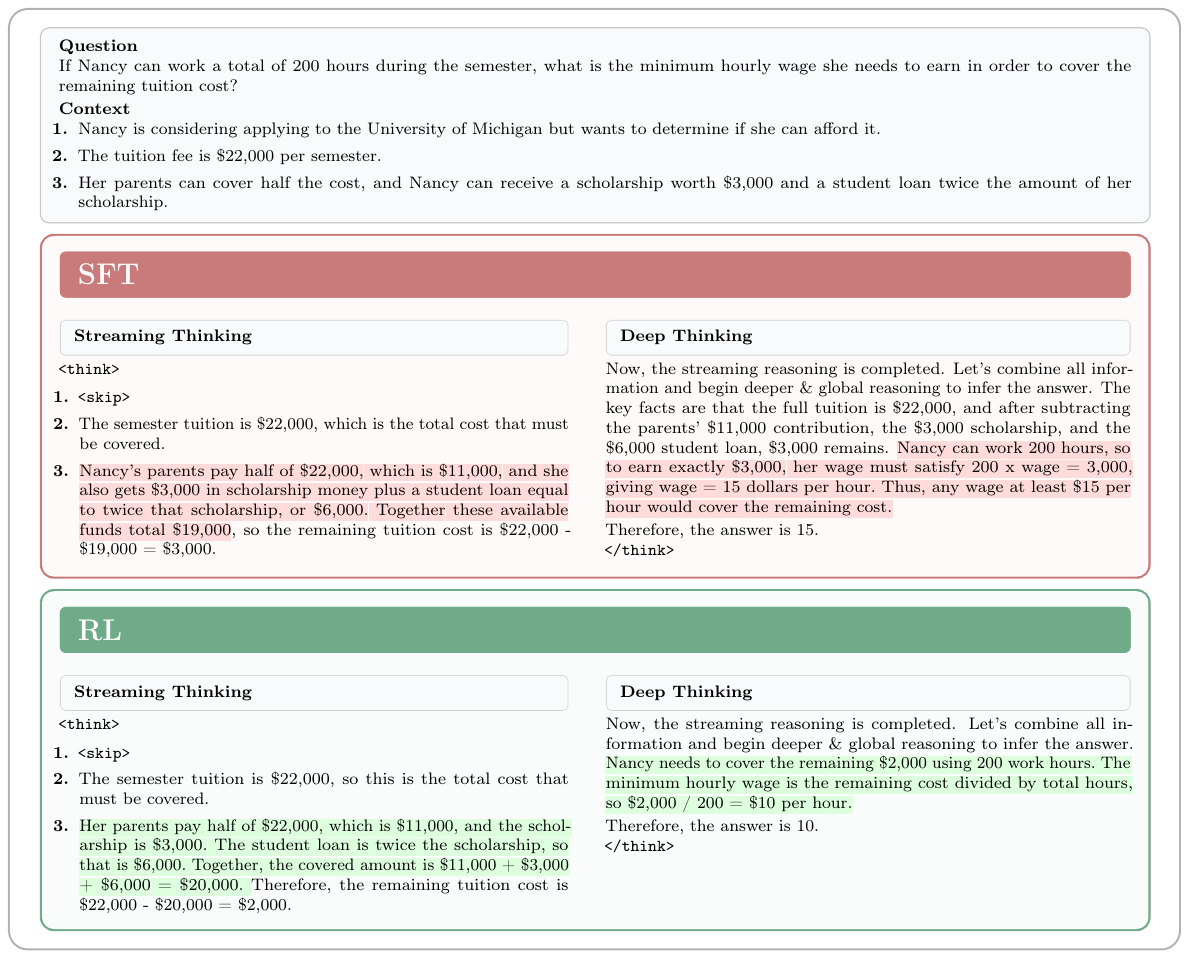}
    \caption{Comparison between SFT and RL on a mathematical reasoning example.}
    \label{fig:sft_rl}
\end{figure*}

The streaming rollout extends vLLM generation from a single full-prompt call to a round-based state machine. For each prompt, the rollout state initially contains only the first revealed source segment and the assistant prefix. At round $t$, the state is converted to a \texttt{TokensPrompt} containing all source segments revealed so far and all previously generated target tokens. The model then decodes a streaming reasoning segment $R_t$ until a streaming stop token such as \texttt{<EOT>} is reached. If unrevealed source segments remain,
the next sentence is appended to the source side and the request is requeued; otherwise the final round uses the deep-reasoning sampling configuration and decodes until the final termination token.

To preserve StreamingThinker's position semantics during vLLM rollout, each
prompt token is annotated with a source or target role, including its segment index. The custom vLLM model-input builder maps these roles to group position IDs: source tokens and target tokens maintain independent position counters that both start from zero. For strict streaming masks, the same segmented roles are also used to build a block-diagonal visibility mask during prefill. Thus, even though vLLM schedules batched generation efficiently, the model observes only the source segments that should be visible at the current streaming step.
This implements the StreamingThinker attention and position design inside the RL rollout engine.

\paragraph{Forward Logits Calculation}
PPO-style updates require the log-probabilities of the sampled response under
the old policy, the reference policy, and the updated actor. A subtle issue is
that the physical rollout tensor layout differs from the logical streaming
layout: prompts are left-padded before rollout, responses are right-padded after
generation, and the target sequence contains both streaming and deep reasoning
segments. Before each actor forward pass, AdaSR therefore removes invalid
padding, concatenates the real source tokens with the real response tokens, and
right-pads the resulting logical sequences within the micro-batch. The
streaming metadata records the source length, target length, and per-segment
boundaries of each sample.

Given this packed representation, the Qwen3 streaming forward pass constructs a
sentence-level streaming causal mask. Streaming tokens are allowed to attend to
past target tokens and to source segments that have already been revealed, but
not to future source segments. The final deep-reasoning segment is allowed to
attend to the full source context and previous streaming thoughts. Logits for
target tokens are sliced from the positions immediately preceding each response
token, and the resulting log-probabilities are written back into the fixed
response-width tensors expected by verl. Consequently, the PPO ratio compares
old and new policy probabilities for exactly the same sampled tokens under the
same streaming attention constraint, while HRPO can assign streaming, deep, and
global advantages to their corresponding token ranges.
\section{Latency Analysis of Streaming Reasoning}
\paragraph{Time Delay}
\label{subsec:latency}
Time delay measures the wall-clock latency between the arrival of the final input token and the generation of the first answer token, directly reflecting the user-perceived waiting time after the input stream ends. As shown in Figure~\ref{fig:latency}, read-then-think defers all reasoning until the full input is observed, so the entire reasoning process contributes to post-input latency. In contrast, streaming thinking performs intermediate reasoning during input reception, amortizing computation over the input stream and reducing residual latency after the final input segment arrives. Following StreamingThinker~\cite{tong2025streamingthinker}, we set the input reading speed to 150 words per minute.

\section{Case Study}
\paragraph{Comparison of SFT and RL for Streaming Paradigm for Math Reasoning}
Figure~\ref{fig:sft_rl} is an extracted example comparing SFT and RL in mathematical reasoning. As shown in the highlighted section, the RL model performs the calculation correctly in the third segment of the streaming thinking process, whereas the SFT model makes an arithmetic error at the same step. This error in the SFT reasoning is then propagated into the  deep thinking stage, ultimately leading to an incorrect answer.

\end{document}